\documentclass[11pt]{article}

 \usepackage[utf8]{inputenc}
 \usepackage[english]{babel}
 \usepackage{cite,wrapfig}
 \usepackage{amsfonts}
 \usepackage{graphicx, epstopdf}
 \usepackage[useregional]{datetime2}
 \usepackage{graphics}
 \usepackage{amsthm, subfigure}
 \usepackage{amssymb}
 \usepackage{amsmath}
 \usepackage{caption}
 \usepackage{bbm}
 \usepackage[section]{placeins}
\usepackage{algorithm}
\usepackage{algpseudocode}
 \usepackage{xcolor}
\usepackage{booktabs}
 \usepackage{array}
\usepackage{verbatim}
\usepackage{float}
\usepackage{placeins}
\usepackage{stackengine}
\usepackage{hyperref}

\usepackage[tablename=TABLE]{caption}
\usepackage[figurename=FIG.]{caption}

\newtheorem{theorem}{Theorem}[section]

\newtheorem{lemma}{Lemma}[section]
\newtheorem{proposition}{Proposition}[section]

\theoremstyle{definition}
\newtheorem{definition}{Definition}[section]

\theoremstyle{definition}

\theoremstyle{remark}
\newtheorem{remark}{Remark}

\def\P{\mathbb{P}}

\newcommand{\R}{\mathbb{R}} 
\newcommand{\N}{\mathbb{N}} 
\newcommand{\W}{\mathbb{W}} 
\newcommand{\D}{\mathcal{D}} 
\newcommand{\Pro}{\mathbb{P}} 
\newcommand{\X}{\mathbb{X}} 
\newcommand{\1}{\mathbbm{1}}

\title{Bayesian Topological Learning for Classifying the Structure of Biological Networks }

 \author{Vasileios Maroulas \thanks{Department of Mathematics, University of Tennessee, Knoxville, TN }
  \and Cassie Putman Micucci  \footnotemark[1]
 \and Farzana Nasrin  \thanks{Department of Mathematics, University of Hawaii at Manoa, Honolulu, HI (\href{mailto:fnasrin@hawaii.edu}{fnasrin@hawaii.edu} )}
 }
\date{}
\begin{document}
\maketitle
\providecommand{\keywords}[1]{\textbf{\textit{Keywords}} #1}

\begin{abstract}
Actin cytoskeleton networks generate local topological signatures due to the natural variations in the number, size, and shape of holes of the networks. Persistent homology is a method that explores these topological properties of data and summarizes them as persistence diagrams. In this work, we analyze and classify these filament networks
 	 by transforming them into persistence diagrams whose variability is quantified  via a Bayesian framework on the space of persistence diagrams. 
	   The proposed generalized Bayesian framework adopts an independent and identically distributed cluster point process characterization of persistence diagrams and relies on a substitution likelihood argument. 
	    This framework provides the flexibility to estimate the posterior cardinality distribution of points in a persistence diagram and the posterior spatial distribution simultaneously.
	    We present a closed form of the posteriors under the assumption of Gaussian mixtures and binomials for prior intensity and cardinality respectively. 
 	    Using this posterior calculation, we implement a Bayes factor algorithm to classify the actin filament networks  and  benchmark it against several state-of-the-art classification methods.

\end{abstract}

\section{Introduction}
The actively functioning transportation of various particles through intracellular movements is a vital process for cells of living organisms (\cite{porter2016filaments}). 
Such transportation must be intricately organized due to the tightly packed nature of the interior of a cell at the molecular level (\cite{breuer2017system}). 
The actin cytoskeleton,  which consists of actin filaments cross-linking with myosin motor proteins along with other pertinent binding proteins,  is an important component in plant cells that determines the structure of the cell and provides transport of cellular components (\cite{freedman2017versatile, breuer2017system}). 
Although researchers have investigated the molecular features of actin cytoskeletons (e.g., \cite{Staiger2000, Shimmen2004,freedman2017versatile,Mlynarczyk2019}), 
the underlying process that determines their structures and how these structures are linked to intracellular transport remains undetermined (\cite{thomas2009actin,madison2013understanding}).
A crucial step to understand this transport is to define quantitative measures of the actin cytoskeleton's  structure, and understand the different structural networks of filaments on which the organelles are moving. However there is not a method for fully depicting the characteristics of networks.

\begin{wrapfigure}{r}{0.4\textwidth}
	\vspace{-15pt}
	\centering{
		\includegraphics[width=2in,height=1.2in]{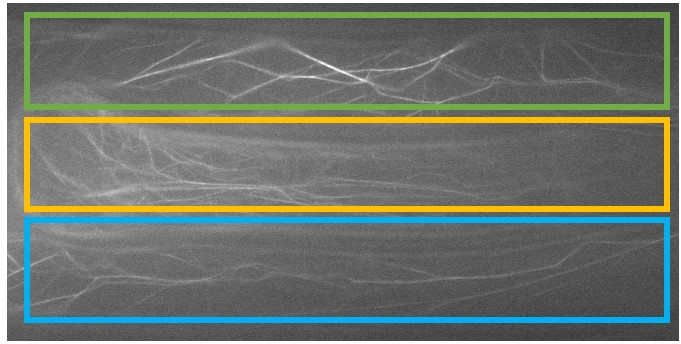}
	}
	\caption{ \footnotesize An electron micrograph of an actin filament. \label{fig:actin}}
\end{wrapfigure}

On the other hand, from a closer look, the inherent variation in size, density, and positioning of actin filaments yields topological signatures in the cytoskeleton's network, (\cite{Tang2014}).
In this article, we develop a fully data-driven Bayesian learning method, which could aid researchers by providing a pathway to predict cytoskeleton structural properties by classifying
actin filament networks to identify the effect of the number of cross-linking proteins on the network. 
Fig. \ref{fig:actin} presents an electron micrograph of actin filaments and highlights several segments of the filament network to show the variation in the number of cross-linking proteins. 
The green segment shows brighter regions than the rest, as this includes thicker actin cables. 
This segment also includes larger holes within the loops created by the actin cables.
This difference in segments reflects the role of binding proteins in linking actin filaments together into bundles and networks.
With more cross-linking proteins available, the cell has networks with many binding locations, which create thicker cables and larger loops within the whole structure of the actin cytoskeleton. 
When viewed through the lens of topology,  these networks show dissimilarity due to the presence and size of loops. 
Differentiating between the empty space and the connectedness of these networks allows us  to  create  an  accurate  classification  rule
using topological methods. Although we focus on our analysis to the classification of actin filament networks, the topological Bayesian framework could be generalized to other data sets.

	Persistent homology is a powerful topological data analysis (TDA) tool that provides a robust way to model the topology of data and summarizes  salient features with persistence diagrams (PDs).
	 These diagrams are multisets of points in the plane, each point representing a homological feature whose time of appearance and disappearance is contained in the coordinates of that point (\cite{Edelsbrunner2010}). Persistent homology has proven to be promising in a variety of applications such as shape analysis \cite{Adcock2016,Patrangenaru2018,Lum2013}, image analysis \cite{Bonis2016,Carriere2015,Carlsson2008, Guo2018}, neuroscience \cite{Chung2015,Sizemore2018,Babichev2017,Bendich2016,Biscio2019,Nasrin2019}, sensor networks \cite{Dlotko2012,Carlsson2010, Carlsson2009,Silva2006, Silva2007}, biology \cite{Sgouralis2017,Marouls2015,Mike2016,Nicolau2011,Gameiro2015,Kusano2016,Ciocanel2019}, dynamical systems \cite{Khasawneh2016,Perea2015,Rouse2015}, action recognition \cite{Venkataraman2016}, signal analysis \cite{Marchese2018,Marchese2016,Pereira2015,Seversky2016}, chemistry and material science, \cite{Xia2015,Lee2017,Ichinomiya2017,Kimura2018,Maroulas2019a, Townsend2020}, and genetics \cite{Humphreys2019,Emmett2014}.

 While there are several methods present in the literature to compute PDs, we choose geometric complexes that are typically used for applications of persistent homology to data analysis; see \cite{ Edelsbrunner2010} and references therein.  
 The homological features in PDs have no intrinsic order, implying that they are sets as opposed to vectors. 
 Due to this, the utilization of PDs in machine learning algorithms is not straightforward. 
 Some researchers map the PDs into Hilbert spaces to adopt traditional machine learning tools (see e.g., \cite{Fabio2015,Turner2014,Adams2017,Bubenik2015,Reininghaus2015}). 
 Direct use of PDs for statistical inference and classification has been developed by several authors such as  \cite{Maroulas2019,Maroulas2019a,Marchese2018,Bobrowski2017,Fasy2014,Mileyko2011,Robinson2017}; and \cite{Bubenik2018}.

In this paper, we quantify the variability of PDs through a novel Bayesian framework by considering PDs as a collection of points distributed on a pertinent domain space,  where the distribution of the number of points is also an important feature. 
This setting leads us to view a PD through the lens of an independent and identically distributed (i.i.d.) cluster point process (PP) (\cite{Daley1988}).  
An i.i.d. cluster PP consists of points that are i.i.d. according to a probability density but have an arbitrary cardinality distribution. 
For example, an i.i.d. cluster PP is reduced to the classical Poisson PP if the points in a PD are spatially distributed according to a Poisson distribution. 
The study in \cite{Maroulas2019a} implicitly estimates the cardinality of a PD by integrating the intensity of a Poisson PP.
The framework of \cite{Maroulas2019a} also yields that the variance is equal to the mean and leads to an estimation of cardinality with high variance whenever the number of points in a PD is high.
However, modeling PDs as i.i.d. cluster PPs allows us to estimate the intensity and the cardinality component of the distribution simultaneously.  This is very critical as the importance of cardinality in PDs has been underlined in problems related to statistics and machine learning \cite{Fasy2014,Kerber2017}.

Our Bayesian framework
quantifies prior uncertainty with given intensity and cardinality for an i.i.d. cluster PP. 
 The likelihood  in our model represents the level of belief that observed diagrams are representative of the entire population and is defined through marked point processes (MPPs). 
A central idea of this paper is to develop posterior distributions of the spatial configuration of points on persistence diagrams and their associated number instead of the point clouds in the data generating space. The persistence diagrams summarize their topology which in turn is employed in the classification algorithm.  By viewing point clouds through their topological descriptors, the proposed framework can reveal essential shape peculiarities latent in the point clouds. Our Bayesian method adopts a substitution likelihood technique by Jeffreys in \cite{Jeffreys1961} instead of considering the full likelihood for the point cloud.
 Due to the nature of PDs, an observed PD contains points that correspond to the latent topology in the underlying data as well as points that solely arise due to noise in the data. 
 Our Bayesian model addresses instances of noise by means of an i.i.d. cluster PP. In particular, we are able to quantify the  uncertainty with an estimated intensity and cardinality using the i.i.d. cluster PP.
 This framework estimates the posterior cardinality and intensity simultaneously, which 
 provides a complete knowledge of the posterior distribution.

 Another key contribution of this paper is the derivation of a closed form of the posterior intensity, which relies on Gaussian mixture densities for prior intensities \emph{and} a closed form for the posterior cardinality, which uses binomial priors. 
 The direct benefits of this closed form solution of the posterior distribution are two-fold: (i) it demonstrates the computational tractability of the proposed Bayesian model and  (ii) it provides a means to 
 develop a robust classification scheme through Bayes' factors. Another  computational benefit of these closed forms is the quantification of the intensity of the unexpected PP by means of an exponential  density. The exponential density is an ideal choice because (i) it gives a natural intuition of the unexpected (noise) features, and (ii) it provides a more computationally automatic approach as we only need to modify one parameter.  
This Bayesian paradigm provides a method for the classification of actin filament networks in plant cells that captures their distinguishing topological features.

Overall, the contributions of this work are:
\begin{enumerate}
\item A generalized Bayesian framework that simultaneously estimates  the spatial and the  cardinality distribution of PDs using i.i.d. cluster PPs. 
        \item A general closed form expression of both the posterior spatial distribution and the posterior cardinality distribution of PDs.
        \item A Bayesian classification algorithm for actin filament networks of plant cells that directly incorporates the variations in topological structures of those networks such as number and size of loops. 
    \end{enumerate}

  This paper is organized as follows. 
  Section \ref{sec:prelm} provides a brief overview of PDs and PPs. 
  In Section
 \ref{sec:main}, we establish the Bayesian framework for PDs and provide the update formulas for intensity and cardinality. 
 Then Subsection \ref{sec:gm_post} introduces a closed form representation of the posterior intensity and cardinality utilizing Gaussian mixture models and binomial distributions respectively. 
 Detailed demonstrations of this closed form estimation are presented in Subsection \ref{subsec:ex_post}.   
 To assess the capability of our Bayesian method, we investigate a problem of classifying filament networks of plant cells in Section \ref{sec:class}. 
 Finally, we end with the conclusion in Section \ref{sec: conclusion}. We delegate all of the proofs,  as well as some definitions, lemmas, and notations required for the proofs to the supplementary
materials.
 
 \vspace{-0.2in}
\section{Preliminaries} \label{sec:prelm}

We begin by discussing the necessary background for generating Bayesian models for PDs. In Subsection \ref{subsec:persistence diagram}, we briefly review simplicial complexes, the building blocks for constructing PDs. 
Pertinent definitions, theorems, and some basic facts about i.i.d. cluster point processes (PPs) are discussed in Subsection \ref{subsec:iid pp} .

\subsection{Persistence Diagrams} \label{subsec:persistence diagram}

\begin{definition}
The convex hull of a finite set of points $\{x_i\}_{i=1}^n$ is given by $\sum_{i=1}^n \alpha_i x_i$, where $\alpha_i \geq 0$ for all $i$ and $\sum_{i=1}^n \alpha_i =1$.
\end{definition}

\begin{definition}
The set of points $\{x_i\}_{i=1}^n$ is affinely independent if whenever $ \sum_{i=1}^n \alpha_i x_i = 0$ and $\sum_{i=1}^n \alpha_i = 0$, then $\alpha_i = 0$ for all $i$.
\end{definition}

\begin{definition}
    A $k$-simplex is the convex hull of an affinely independent point set of cardinality $k+1$. The convex hull of a nonempty subset of the $k$ points in a $k+1$ simplex is called a face of a simplex.
    \end{definition}
    
    \begin{definition}
      A simplicial complex $\sigma$ is a collection of simplices such that for
        every set $A$ in $\sigma$ and every nonempty set $B \subset A$, we have that $B$ is in $\sigma$.
        \end{definition}


        \begin{definition}
      The Vietoris-Rips complex for threshold $\epsilon > 0$, denoted VR$(\epsilon)$, is the abstract simplicial complex determined in the following way: a $k$-simplex with vertices given by $ k+1$ points in $X$ is included in $VR(\epsilon)$ whenever $\epsilon/2$ balls placed at the points all have pairwise intersections.  
\end{definition}

\begin{figure}
    \centering
    \subfigure[]{\includegraphics[width=1.2in, height = 1.6in]{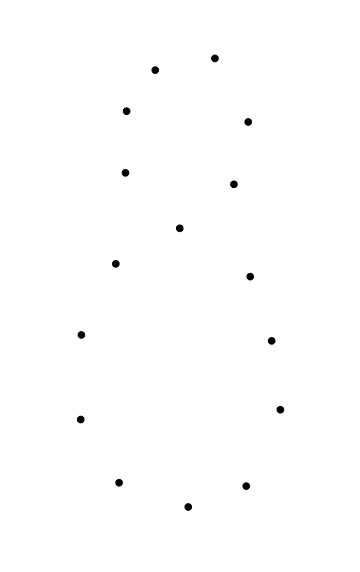}}\hspace{0.1in}
    \subfigure[]{\includegraphics[width=1.2in, height = 1.6in]{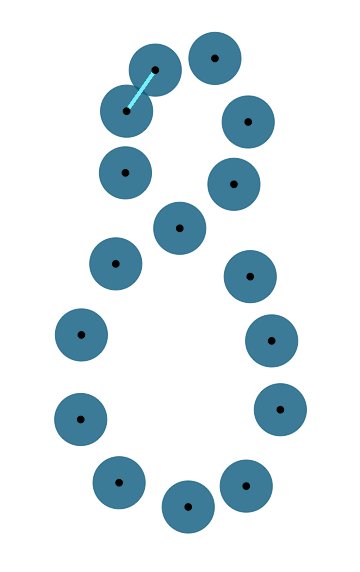}}\hspace{0.1in}
    \subfigure[]{\includegraphics[width=1.2in, height = 1.6in]{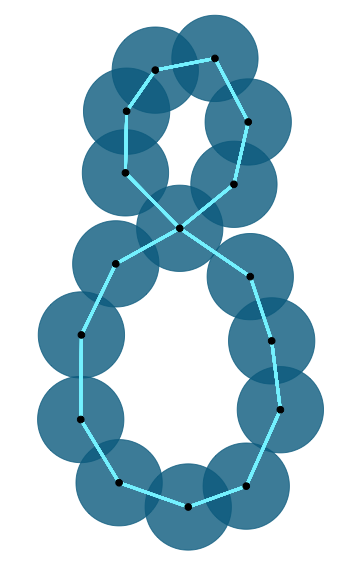}}\hspace{0.1in}
    \subfigure[]{\includegraphics[width=1.9in, height = 1.6in]{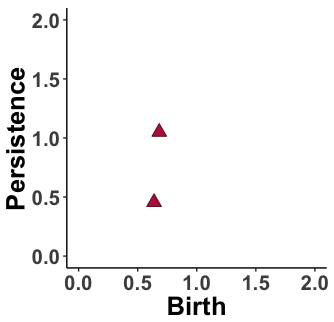}}\hspace{0.1in}
    \caption{{(a) An underlying dataset of points. 
    (b) The Vietoris-Rips complex consisting of the points and the light blue line segment. 
    (c) The Vietoris-Rips complex consisting of the points and line segments that now form a figure-eight, which has two 1-dimensional holes.
    (d) The tilted PD for dimension 1 for (a) has two points.
    }}
        \label{fig:VR}
        
\end{figure}

Formally, for each homological dimension, a PD is a multiset of points $(b,d)$, where $b$ is the radius in the Vietoris-Rips complex at which a homological feature is born and $d$ is the radius at which it dies. 
To facilitate visualization and preserve the geometric information, we apply the linear transformation $(b,p)=T(b,d) = (b,d-b)$ to each point in the diagram. We refer to the resulting coordinates as birth ($b$) and persistence ($p$), respectively, in  $\mathbb{W} := \{(b,p) \in \R^{2} | \,\, b,p \geq 0\}$ and call this transformed PD a tilted representation (Fig. \ref{fig:VR} (d)). 
Hereafter whenever we refer to PDs, we imply their tilted representations. 
 Intuitively, the homological features represented in a PD are connected components or holes of different dimensions.
 For example, a 0-dimensional homological feature is a connected component, a 1-dimensional feature is a loop, and a 2-dimensional feature is a void.
 An example of the evolution of the Vietoris-Rips complex and a corresponding PD is given in Fig. \ref{fig:VR}.
 
 \subsection{I.I.D. Cluster  Point Processes} \label{subsec:iid pp}
This section contains basic definitions and fundamental theorems related to i.i.d. cluster PPs.
Detailed treatments of i.i.d. cluster PPs can be found in \cite{Daley1988} and references therein. 
Throughout this section, we let  $\X$ be a Polish space and $\mathcal{X}$ be its Borel $\sigma-$algebra.

\begin{definition}
    A finite point process $(\{\rho_n\},\{\P_n (\bullet)\})$ consists of a cardinality distribution $\rho_n$ with $\sum_{n=0}^{\infty} \rho_n = 1$ and a symmetric probability measure $\P_n$ on $\mathcal{X}^n$, where $\mathcal{X}^0$ is the trivial $\sigma$-algebra.
    \label{def:fpp}
\end{definition}

 To sample from a PP,
first one draws an integer $n$ from the cardinality distribution $\rho_n$.
Then the $n$ points $(x_1,\dots,x_n)$ are spatially distributed according to a draw from $\P_n$.
Since PPs model unordered collections of points, we need to ensure that $\P_n$ assigns equal weights to all $n\,!$ permutations of $(x_1,\dots,x_n)$.
The requirement in Def. \ref{def:fpp} that $\P_n$ is symmetric guarantees this. 
A natural way to work with random collections of points is the Janossy measure, which combines the cardinality and spatial distributions, while disregarding the order of the points. 

\begin{definition}
 For disjoint rectangles $A_1, \ldots, A_n$,  the Janossy measure for a finite point process is given by
   $ \mathbb{J}_n (A_1 \times \cdots \times A_n) = n! \rho_n \P_n (A_1 \times \cdots \times A_n). $   
\end{definition}


\begin{definition} \label{def:iid cluster}

	An i.i.d. cluster PP $\Psi$ is a finite PP on the space $(\mathbb{X},\mathcal{X})$ which has points that: 
	(i) are located in $\X=\R^d$, 
	(ii) have a cardinality distribution $\rho_n$ with $\sum_{n=0}^{\infty} \rho_n=1$, and 
	(iii) are distributed according to some common probability measure $F(\cdot)$ on the Borel set $\mathcal{X}$. 
\end{definition}

We consider Janossy measures for the point process $\Psi$, $\mathbb{J}_n^{\Psi}$, that admit densities $j_n$ with respect to a reference measure on $\mathbb{X}$ due to their intuitive interpretation. In particular, for an i.i.d. cluster PP $\Psi$, if $F(A) = \int_A f(x) dx$ for any $A \in \mathcal{X}^n$, then $j_n (x_1, \ldots, x_n) = \rho_n n! f(x_1) \cdots f(x_n)$ determines the probability density of finding the $n$ points at their respective locations according to $F$. 
The $n!$ term gives the number of ways the points could be at these positions. 
For a finite intensity measure $\Lambda$ on $\mathbb{X}$ that admits the density $\lambda$, 
 we also have $f(x) =\frac{\lambda(x)}{\Lambda(\mathbb{X})}$.
The intensity is the point process analog of the first order moment of a random variable. 
Precisely, the intensity density $\lambda(x)$ is the density of the expected number of points per unit volume at $x$. 
Hereafter, we sufficiently characterize our i.i.d. cluster PPs with intensity and cardinality measures. 
Next, we define the marked PP, which provides a formulation for the likelihood model used in our Bayesian setting.
 Let $\mathcal{M}$ be a Polish space that represents the mark space, and let its Borel $\sigma-$ algebra be $\mathcal{M}$.

\begin{definition}
	\label{def:marked_iid cluster_process}
	
A marked i.i.d. cluster PP $(\Psi,\Psi_{M})$ is a finite PP on $\mathbb{X} \times \mathbb{M}$ such that: 
(i) $\Psi=\left(\left\{\rho_{n}\right\},\left\{\Pro_{n}(\bullet)\right\}\right)$ is an i.i.d. cluster PP on $\mathbb{X}$, and 
(ii) for a realization  $(\mathbf{x},\mathbf{m}) \in \mathbb{X} \times \mathbb{M}$, the marks $m_i$  of each $x_i \in \mathbf{x}$ are drawn independently from a given stochastic kernel $\ell(\bullet|x_i)$.
\end{definition}

\begin{remark}
\label{remark:marked_iid} A marked point process $(\Psi,\Psi_{M})$ is a bivariate PP where one point process is
parameterized by the other. 
Therefore, if the cardinalities of $\mathbf{x}$ and $\mathbf{m}$ are equal, then the conditional density for $\mathbf{m}$ is $\ell(\mathbf{m}|\textbf{x}) = \frac{1}{n!} \sum_{\pi \in \mathcal{S}_{n}}\prod_{i=1}^{n}\ell(m_{i}|x_{\pi(i)})$, where $\mathcal{S}_n$ is the set of all permutations of $(1,\dots,n)$.
Otherwise, the density can be taken as 0.

\end{remark}

\section{Bayesian Inference} \label{sec:main}

Considering a sample PD from an i.i.d. cluster process,
we define a generalized Bayes' theorem for PDs. 
First, we consider the underlying prior uncertainty of a PD $D_X$ as generated by an i.i.d. cluster PP $\D_X$ with intensity $\lambda_{\D_X}$ and cardinality $\rho_{\D_X}$.
The cardinality distribution $\rho_{\D_X}(n)$ is defined as the probability, $P(|\D_X|=n)$, of the number of elements in the PP $\D_X$ to be $n$, where $|\cdot|$ denotes the cardinality of a PD. Due to the nature of PDs, we may encounter two scenarios for any point $x$ in $\D_X$. We accommodate these two possibilities by means of a probability function $\alpha(x)$. In particular, the
scenario of observing $x$ happens with probability  $\alpha(x)$, and otherwise with  probability  $1-\alpha(x)$.

\begin{figure}[h!]
    \centering

    \subfigure[]{\includegraphics[width=1.45in,height=1.2in]{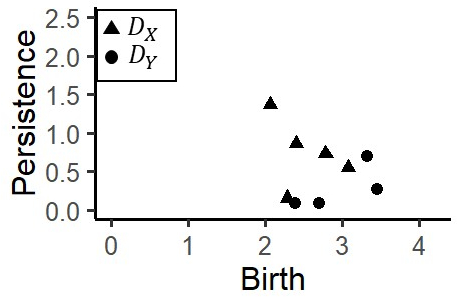}}
    \subfigure[]{\includegraphics[width=1.45in,height=1.2in]{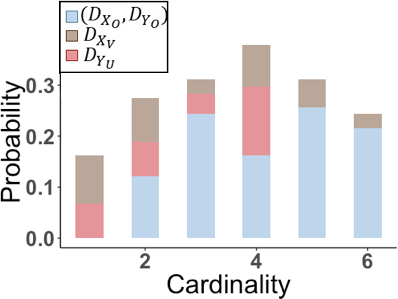}}
    \subfigure[]{\includegraphics[width=1.45in,height=1.2in]{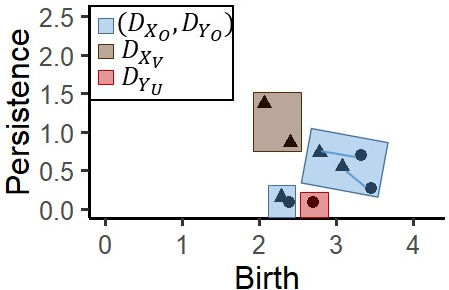}}
     \subfigure[]{\includegraphics[width=1.45in,height=1.2in]{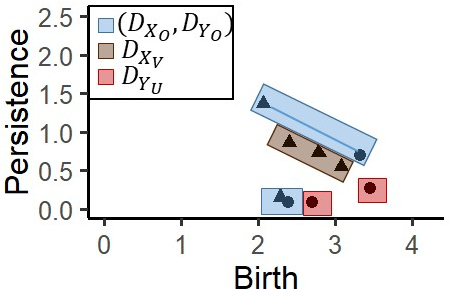}}

    \caption{  (a) A sample $D_X$ (triangles) from the prior PP $\mathcal{D}_X$ and an observed PD $D_Y$ (dots) generated from $\mathcal{D}_Y$. (b)  is an example of the cardinality distribution of prior and observed PDs which  shows some  possible configurations of the cardinality probability of $\D_{X_O},\D_{X_V},\D_{Y_O}$, and $\D_{Y_U}$.
    (c-d) are similar color-coded 
    representations of possible configurations for the observed and vanished features in the prior, along with the marked PP $(\D_{X_O},\D_{Y_O})$ and the unexpected features of the data $\D_{Y_U}$.} 
    \label{fig:dxdy}
\end{figure}


Next, we establish the likelihood model by employing the theory of marked PPs.
For a marked PP, the intensity (spatial) joint probability density is computed using a stochastic kernel. More specifically, $(\mathbf{x},\mathbf{m}) \in \X \times \X_M$ implies that the points $m(x_i) \in \mathbf{m}$ are marks of $x_i \in \mathbf{x}$ that are drawn independently from a stochastic kernel $\ell: \mathbb{X} \times \X_M \rightarrow \R^{+}\cup \{0\}$. This kernel, in turn, provides the conditional density $\ell(\mathbf{m}|\mathbf{x})$, which is nonzero only if the cardinalities of $\mathbf{x}$ and $\mathbf{m}$ are equal as discussed in Remark  \ref{remark:marked_iid}. On the other hand, the cardinality likelihood is obtained by the conditional distribution of the observed PD $D_Y$ given that there are $n$ points in $\D_X$. This cardinality likelihood incorporates the prior intensity and cardinality distribution, the stochastic kernel $\ell$, and the distribution of the topological noise in the observed data. 
In order to fully describe the structure of PDs, we must define one last PP that models the topological noise in the observed data. 
Intuitively, this PP consists of the points in the observed diagram that fail to associate with the prior. We define this as an i.i.d. cluster PP $\D_{Y_U}$ with intensity $\lambda_{\D_{Y_U}}$ and cardinality $\rho_{\D_{Y_U}}$.

Fig. \ref{fig:dxdy}  gives a visual representation to illustrate the contribution of the prior and the observations to the spatial and the cardinality distributions of the Bayesian framework.  
For this we superimpose two PDs: one is a sample from the prior and the other is the observed PD (see Fig. \ref{fig:dxdy} (a)). 
Any point $x$ in $\D_X$ is equipped with a probability of being observed and not being observed, which we present using blue and brown respectively in Fig. \ref{fig:dxdy} and denote them as $D_{X_O}$ (observed) and $D_{X_V}$ (vanished), respectively. 
Presumably, any point $x \in D_{X_O}$ has an association with one feature $y$ in the observed PD $D_Y$, and we call those features in $D_Y$ as $D_{Y_O}$.
This implies that for any possible configuration, the number of points in $D_{X_O}$ will be the same as $D_{Y_O}$ and we use blue to represent that. 
Also, there can be features in the observed PD $D_Y$ that are generated from noise or unanticipated geometry in the prior $\mathcal{D}_{X}$, and we denote them as $D_{Y_U}$ and call them unexpected features (presented as red in Fig. \ref{fig:dxdy}).  
Fig. \ref{fig:dxdy} (b) shows different possible scenarios for the relationship of the prior to the data likelihood for the cardinality distribution. 
As any point in $D_{X_V}$ has no association with  features in the observed PD $D_Y$, the case of having all the points in $D_{X_V}$ indicates that in the observed PD we encounter only the unexpected $D_{Y_U}$ (the first bar in Fig. \ref{fig:dxdy} (b)). 
As some points of $\D_Y$ are more likely to be marks than others, we illustrate these instances with different levels for the blue parts of the cardinality bars. 
The last two bars demonstrate the cases where all of the points in $\D_Y$ are expected to be marks of the prior features; this is encountered in the presence of very low noise in data.

Fig. \ref{fig:dxdy} (c)  and (d) show different possible scenarios for the relationship of the prior to the data likelihood for the spatial distribution.
We use boxes of corresponding colors to highlight the decomposition of the PP $\D_X$  into  $D_{X_O}$ and $D_{X_V}$, and $\D_Y$ into $D_{Y_O}$  and $D_{Y_U}$. 
For example, the observed $D_{X_O}$ and vanished $D_{X_V}$ features are presented as triangles inside of blue and brown boxes respectively in Fig. \ref{fig:dxdy} (c) and (d).
All associations between points $D_{X_O}$ and $D_{Y_O}$ together constitute the marked PP which admits a stochastic kernel $\ell(y|x)$. 
This indicates that the point $x$ may have any point $y \in D_Y$ as its mark, but intuitively some marks should be more likely than others.  
In  Fig. \ref{fig:dxdy} (c) and (d) we give examples of these different associations, which are indicated by pairs inside of the blue box. Finally, the unexpected features in $D_{Y_U}$ are presented as  dots inside of red boxes  in Fig. \ref{fig:dxdy} (c) and (d).
Finally, the posterior 
intensity and cardinality are given in the theorem below, whose proof is delegated to Section \ref{sup_proof} in the supplementary materials.

\begin{theorem}  \label{thm:bayes}
For a random PD, denote the prior intensity and cardinality by $\lambda_{D_X}$ and $\rho_{\D_X}$, respectively. 
Suppose $\alpha(x)$ is the probability of observing a prior feature, and 
$\mathcal{D}_{X_O}$ and $\mathcal{D}_{X_V}$ are two instances of observed and vanished features in the prior respectively. If $\ell(y|x)$ is the stochastic kernel that links $\D_{Y_O}$ with $\D_{X_O}$, and  $\lambda_{\D_{Y_U}}$ and $\rho_{\D_{Y_U}}$ are the intensity  and cardinality of $\D_{Y_U}$ respectively,  then for a set of independent samples of PDs $ D_{Y_{1:m}}=\{ D_{Y_1},\cdots,D_{Y_m} \}$ from $\D_Y$ with cardinalities $K_1, \cdots, K_m$, we have the following posterior intensity and cardinality:

 \begin{equation}\label{post intensity_1}
\small\lambda_{\D_X|D_{Y_{1:m}}}(x)  = \frac{1}{m}\sum_{i=1}^m \Bigg[(1-\alpha(x)) \lambda_{\D_X}(x) B(\emptyset) + \sum_{y \in D_{Y_i}}\frac{\alpha(x)\ell(y|x)\lambda_{\mathcal{D}_X}(x) B(y)}{\lambda_{\D_{Y_U}}(y)}  \Bigg],\,\,\, \text{and}
\end{equation}

	\begin{equation} \label{post cardinality}
\small\rho_{\D_X|D_{Y_{1:m}}}(n)  =\frac{1}{m}\sum_{i=1}^m\frac{\rho_{\D_X}(n)\Bigg(\!\! \sum_{k=0}^{K_i}  (K_i-k)!\,P_{k}^{n}\,\rho_{\mathcal{D}_{Y_U}}(K_i-k)\,\,(\lambda_{\mathcal{D}_X}[1-\alpha])^{n-k} \,\,e_{K_i,k} (D_{Y_i}) \Bigg)}{\langle \rho_{\D_X},\Gamma_{D_{Y_i}}^{0,0}\rangle},
	\end{equation}
	
 \begin{align}
\text{where}\,\,\small B(\emptyset) &= \frac{\langle \rho_{\D_X},\Gamma_{D_{Y_i}}^{0,1}\rangle}{\langle \rho_{\D_X},\Gamma_{D_{Y_i}}^{0,0}\rangle}, \,\,\,  B(y) = \frac{\langle \rho_{\D_X},\Gamma_{D_{Y_i} \setminus y}^{1,1}\rangle}{\langle \rho_{\D_X},\Gamma_{D_{Y_i}}^{0,0}\rangle}, \,\,\, e_{K_i,k} (D_{Y_i}) = \!\!\!\!\!\sum_{\substack{S_{Y_i} \subseteq D_{Y_i}\\ |S_{Y_i}| = k}} \prod_{y \in S_{Y_i}}
     \,\,\, \frac{\lambda_{\mathcal{D}_X}[ \alpha \ell (y|x)]}{ \lambda_{\mathcal{D}_{Y_U}}(y)},\nonumber\\
  \small   \Gamma_{D_{Y_i}}^{a,b}(\tau)\!\! &= \!\!  \Bigg(\!\! \sum_{k=0}^{\min\{K_i-a,\tau\}}\!\! \!\!\!\! \!\! \!\!  (K_i-k-a)!\,P_{k+b}^{\tau}\,\rho_{\mathcal{D}_{Y_U}}(K_i-k-a)(\lambda_{\mathcal{D}_X}[1-\alpha])^{\tau-k-b}e_{K_i-a,k}(D_{Y_i}) \Bigg), \label{eq:gamma}
 \end{align}
  $f[\zeta] = \int_{\mathcal{X}}\zeta(x)f(x)dx $ is a linear functional, $P_i^n$ is the permutation  coefficient, and the sum in $\Gamma_{D_{Y_i}}^{0,0}(n)$ of Eqn. \eqref{post cardinality} goes from $0$ to $K_i$. 
\end{theorem}

In the posterior intensity expression given in Eqn. \eqref{post intensity_1}, the two terms reflect the decomposition of the prior intensity. Due to the arbitrary cardinality distribution assumption for i.i.d. cluster point processes, the two terms are also weighted by two factors $B(\emptyset)$ and $B(y)$ respectively. 
The first term is for the vanished features $\D_{X_V}$, where the intensity is weighted by $1-\alpha(x)$ and $B(\emptyset)$. The factor $B(\emptyset)$ is encountered since there is no $y \in D_{Y_i}$ to represent the vanished features $\D_{X_V}$. 
The second term in  Eqn. \eqref{post intensity_1} corresponds to the observed part $\D_{X_O}$ and is weighted by $\alpha(x)$ and $B(y)$. The factor $B(y)$ depends on specific $y \in D_{Y_i}$ to account for the associations between the features in $\D_{X_O}$ and those in $D_{Y_i}$.
To be more precise, if $x \in D_X$ is observed, it can be associated with any of the $y \in D_{Y_i}$ and the  remaining points of $D_{Y_i}$, defined as $D_{Y_i} \setminus y$, are considered to either be observed from the rest of the features in $\D_X$ or originated as unexpected features $\D_{Y_U}$.

The posterior cardinality is given in Eqn. \eqref{post cardinality}.
The associated likelihood is given as the sum from
$k=0$ to $K_i$, where $K_i$ is the number of features in $D_{Y_i}$. 
This provides the likelihood of each observed PD $D_{Y_i}$ given that there are $n$ points in $\D_X$. 
In particular, for $k=0$, the cardinality term for the unexpected feature  reduces to $\rho_{\D_{Y_U}}(K_i)$ and the intensity term for the vanished feature reduces to $(\lambda_{\D_X}[1-\alpha])^n$. 
This implies that if the observed PD consists only of unexpected features then all of the points in the prior are most likely to have vanished. 
As the value of $k$ increases, contributions from the unexpected features and vanished features decrease, indicating the presence of more associations between prior and observed features through the marked point process $(\D_{X_O}, \D_{Y_O})$.


\subsection{Closed Form of Posterior Estimation} \label{sec:gm_post}

Next, we present a closed form solution  to the posterior intensity and cardinality equation of Thm. \ref{thm:bayes} by considering  a Gaussian mixture density for the prior intensity and a binomial distribution for the prior cardinality.
Below we specify the necessary components of Thm. \ref{thm:bayes} to derive these closed forms. 

 \noindent \textbf{(M1)} The expressions for the prior intensity $\lambda_{\mathcal{D}_X}$ and cardinality $\rho_{\mathcal{D}_X}$  are:

\begin{equation} \label{eqn:intensity and cardinality of Dx}
 	\lambda_{\mathcal{D}_X}(x) = \sum_{l = 1}^{N}c^{\mathcal{D}_X}_{l}\mathcal{N}^{*}(x;\mu^{\mathcal{D}_X}_{l},\sigma^{\mathcal{D}_X}_{l}\mathbf{I}), \,\,\, \text{and} \,\,\,
 	 \rho_{\mathcal{D}_X}(n) = \binom{N_0}{n} \, \rho_x^n(1-\rho_x)^{N_o-n}, 
 	 \end{equation}
where  
$N$ is the number of components, $\mu^{\D_X}$ is the mean, and $\sigma^{\D_X}\mathbf{I}$ is the covariance matrix of the Gaussian mixture. 
Since PDs are modeled as point processes on the space 
 $\mathbb{W}$ not on $\mathbb{R}^2$, the Gaussian densities are restricted to $\mathbb{W}$ as
	$\mathcal{N}^{*}(z;\upsilon,\sigma \mathbf{I}) := \mathcal{N}(z;\upsilon,\sigma \mathbf{I})\1_{\mathbb{W}}(z),$
	 with mean \textit{v} and covariance matrix $\sigma \mathbf{I}$, and
	$\1_{\mathbb{W}}$ is the indicator function of  $\mathbb{W}$. 
$N_0 \in \mathbb{N}$ is the maximum number of points in the prior PP  and $\rho_x \in [0,1]$ is the probability of one point to fall in the space $\mathbb{W}$.  

 \noindent \textbf{(M2)} The likelihood function $\ell(y|x)$, which is the stochastic kernel of the marked i.i.d. cluster PP $(\D_{X_O},\D_{Y_O})$, takes the form 

\begin{equation} \label{eqn:stachastic kernel gaussian}
 	\ell(y|x) = \mathcal{N}^{*}(y;x,\sigma^{\mathcal{D}_{Y_O}}\mathbf{I}),
 	\end{equation}
where $\sigma^{\mathcal{D}_{Y_O}}$ is the covariance coefficient that quantifies the level of confidence in the observations.

 \noindent \textbf{(M3)} The i.i.d. cluster PP $\D_{Y_U}$, consisting of the unexpected features in the observation, has intensity $\lambda_{\mathcal{D}_{Y_U}}$ and cardinality $\rho_{\mathcal{D}_{Y_U}}$. 
The intensity for $\D_{Y_U}$ takes the form
\begin{equation}
\label{eqn:intensity of Dys Gauss}
 		\lambda_{\mathcal{D}_{Y_U}}(y_\textrm{birth},y_\textrm{pers}) =         \mu_{\mathcal{D}_{Y_U}}^2 \exp({-\mu_{\mathcal{D}_{Y_U}} (y_{\textrm{birth}} +y_{\textrm{pers}}) }).
		\end{equation}

$\mu_{\mathcal{D}_{Y_U}}$ controls the rate of decay away from the origin.
This distribution for $\lambda_{\mathcal{D}_{Y_U}}$ considers points closer to the origin more likely to be unexpected features. 
Points close to the origin in PDs are often created either from the spacing between the point clouds due to sampling or from the presence of noise in the data. 
Typically points with higher persistence or higher birth represent significant topological signatures, so for our analysis we count them as less likely to be unexpected.
The cardinality distribution is
\begin{equation}
\label{eqn:cardinality of Dys}
  		\rho_{\mathcal{D}_{Y_U}}(n) = \binom{M_0}{n} \, \rho_y^n(1-\rho_y)^{M_0-n},   
\end{equation}
where $M_0 \in \mathbb{N}$ is the maximum number of points in the PP $\D_{Y_U}$ and $\rho_y \in [0,1]$ is the probability of one point to fall in the space $\mathbb{W}$.



 \begin{proposition}	\label{prop:post}
 Suppose that $\lambda_{\D_X},  \rho_{\mathcal{D}_X}$,  $\ell(y|x)$, $\lambda_{\mathcal{D}_{Y_U}}$, and  $\rho_{\mathcal{D}_{Y_U}}$  satisfy the assumptions (M1)--(M3), and  $\alpha$ is fixed. Then the posterior intensity and cardinality of Thm. \ref{thm:bayes} are given by:
 
    \begin{align}
\lambda_{\D_X|D_{Y_{1:m}}}(x) &= \frac{1}{m}\sum_{i=1}^m \Bigg[ (1-\alpha) \lambda_{\D_X}(x) B(\emptyset)+ \sum_{y \in D_{Y_i}}\sum_{l=1}^{N} C_{l}^{x|y}\mathcal{N}^*(x;\mu_{l}^{x|y},\sigma_{l}^{x|y}\mathbf{I}) \Bigg] \,\,\, \text{and}\label{post intensity_2}\\
\rho_{\D_X|D_{Y_{1:m}}}(n)&=
\frac{1}{m}\sum_{i=1}^m\frac{\rho_{\D_X}(n)\Gamma_{D_{Y_i}}^{0,0}(n)}{\langle \rho_{\D_X},\Gamma_{D_{Y_i}}^{0,0}\rangle} \label{post cardinality_1},
	\end{align}
where $ \Gamma_{D_{Y_i}}^{a,b}(\tau)$, $B(\emptyset)$, and $B(y)$ are as in Thm. \ref{thm:bayes} with 

\vspace{-0.3in}
 \begin{align}
    e_{K_i,k} (D_{Y_i}) &=\!\!\!\!\!\!\!\!\!\!\!\!\sum_{S_{Y_i} \subseteq D_{Y_i}, |S_{Y_i}| = k} \prod_{y \in S_{Y_i}}
     \,\,\, \frac{\alpha \langle c^{\mathcal{D}_X}, q (y) \rangle}{ \lambda_{\mathcal{D}_{Y_U}}(y)}; \,\,\,q_l (y) = \mathcal{N}(y;\mu_{l}^{\mathcal{D}_X},(\sigma^{\mathcal{D}_{Y_O}}+\sigma_{l}^{\mathcal{D}_X})\mathbf{I});\nonumber\\
      C_{l}^{x|y} &= \frac{ B(y)\,\,\alpha c_l^{\D_X}q_l(y)}{ \lambda_{D_{Y_U}}(y)}; \,\,\, \mu_{l}^{x|y} = \frac{\sigma_{l}^{\mathcal{D}_X} y+\sigma^{\mathcal{D}_{Y_O}}\mu_{l}^{\mathcal{D}_X}}{\sigma_{l}^{\mathcal{D}_X}+\sigma^{\mathcal{D}_{Y_O}}}; \,\,\,\,\, \text{and}\,\,\,\,	\sigma_{l}^{x|y}= \frac{\sigma^{\mathcal{D}_{Y_O}}\,\sigma_{l}^{\mathcal{D}_X}}{\sigma_{l}^{\mathcal{D}_X}+\sigma^{\mathcal{D}_{Y_O}}}. \nonumber
 \end{align}
 \end{proposition}
 
 We present the proof in 
Section \ref{sup_prop_proof} of the supplementary materials. 
One can see that the intensity estimation in Eqn. \eqref{post intensity_2} is in the form of a Gaussian mixture, and hence it is obtained from a conjugate family of priors. However, we do not observe a similar property for the cardinality estimation.
A detailed example of these estimations is provided in Section \ref{subsec:ex_post}. 
The cardinality distribution in Eqn. \eqref{post cardinality_1}  is computed for infinitely many values of $n$, which is unattainable. 
Hence, for the practical application, we must truncate $n$ at some $N_{max}$ such that $N_{max}$ is sufficiently larger than the number of points in the prior PP.
Without loss of generality, we can choose $N_{max} = N_0$.

\subsection{Sensitivity Analysis} \label{subsec:ex_post}
\begin{figure}[h!]
    \centering
 \includegraphics[width=6.5in,height=5.5in]{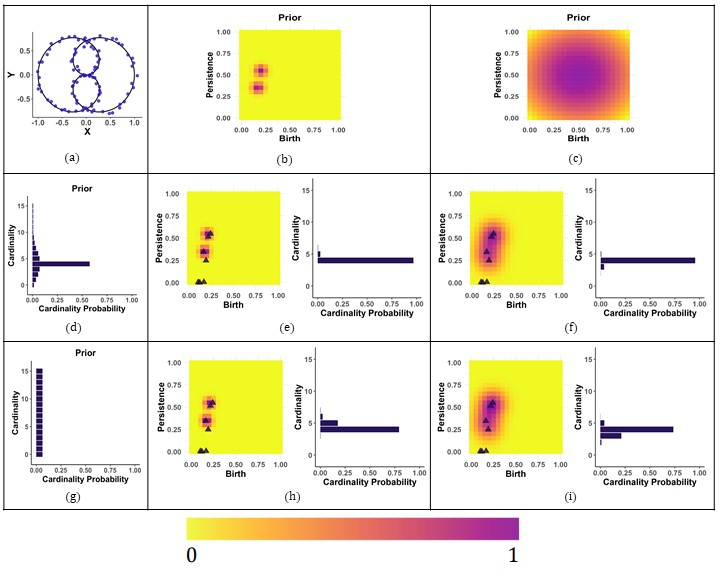}
 
 \vspace{-0.1in}
  \caption{\footnotesize{Posterior intensities and cardinalities obtained for Case-1 by using Proposition \ref{prop:post}. 
  }}
    \label{table:example_post_1}
    \vspace{-0.2in}
\end{figure}

We present the following example to (i) illustrate the estimation of the posterior using Eqns. \eqref{post intensity_2} and \eqref{post cardinality_1}, and (ii) examine the effects on the choice of prior intensity and prior cardinality on the posterior distributions.
To reproduce these results, the interested reader may download our R-package \href{https://github.com/maroulaslab/BayesTDA}{BayesTDA}.
We consider point clouds generated from a polar curve that contains two inner loops (see Fig. \ref{table:example_post_1} (a)) and focus on 1-dimensional features in their corresponding PDs as they are the important homological features of this shape.

\begin{table}[h!]
	\begin{center}
		\begin{tabular}  {|l|c|c|c|c|} 
		\hline 	
		\multicolumn{5}{|c|}{\textbf{Parameters for (M1)}}  \\ 
		\hline
		Prior  & $\mu_{i}^{\mathcal{D}_X}$ &  $\sigma_{i}^{\mathcal{D}_X}$ & $c_{i}^{\mathcal{D}_X}$& $N_0$   \\
		\hline
	\raisebox{0.1in}{Informative}  & \shortstack{$(0.2,0.55)$ \\$(0.17,0.35)$}   &  \shortstack{0.0018 \\ 0.0018} &  \shortstack{2 \\ 2} &  \raisebox{0.1in}{15}  \\
		\hline
	Unimodal Uninformative & $(0.5,0.5)$  & 0.5   &1  & 15  \\ 
		\hline
	\end{tabular}
	
	\caption{ List of parameters for (M1). We take into account two types of prior intensities and cardinalities: (i) informative and (ii) uninformative.} \label{table:prior parameters3}
	\end{center}
	\vspace{-0.3in}
	\end{table}

\begin{table}[h!]
	\begin{center}
	\begin{tabular}  {|c|c|c||c|c|} 
		\hline 	
\textbf{Cases}	& &	\multicolumn{1}{|c||}{\textbf{Parameters for (M2)}} & \multicolumn{2}{|c|}{\textbf{Parameters for (M3)}}  \\ 
		 & & $\sigma^{\mathcal{D}_{Y_O}}$ & $\mu^{\mathcal{D}_{Y_U}}$ &   $\rho_y$   \\
		\hline
 \textbf{Case-1} & (e),(f),(h),(i) & 0.01  & 20   & $0.5$ \\ 
		\hline
	\raisebox{0.2in}{\textbf{Case-2}} & \shortstack{(e),(f),(h),(i) \\ (k) \\ (l)}   & \shortstack{0.01\\0.001\\0.001}   &    \shortstack{20 \\ 25 \\ 16}&  	\raisebox{0.2in}{$0.5$}  \\ 
	\hline
	 \raisebox{0.2in}{\textbf{Case-3}} & \shortstack{(e),(f),(h),(i) \\ (k)-(l) }   & \shortstack{0.01\\0.001}    &    20 &  \shortstack{$0.5$ \\$0.6$} \\ 
	\hline
	\end{tabular}
	\end{center}
	
	\caption{ List of parameters for  (M2) and (M3).  For Case-1, Case-2, and Case-3, we consider the 1-dimensional persistence features obtained from the point clouds  sampled from the polar curve and perturbed by Gaussian noise having variances $0.001I_2$, $0.005I_2$, and $0.01I_2$ respectively.}
	\label{table:prior parameters2}
	\vspace{-0.2in}
\end{table}

\vspace{0.1in}
\begin{wrapfigure}{r}{0.3\textwidth}
\vspace{-15pt}
	\centering{
		\includegraphics[width=1.5in,height=1.5in]{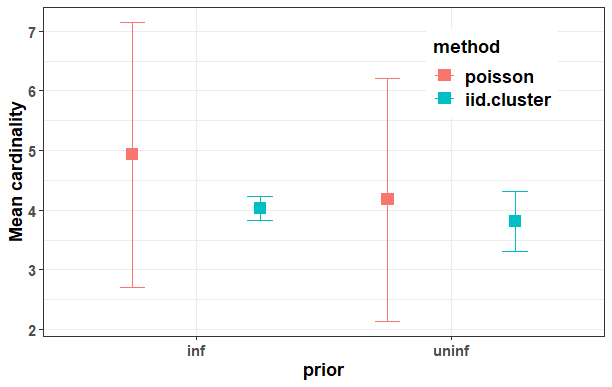}
	}
	\caption{ Cardinality statistics for the posterior cardinalities obtained by using the parameters in Case-1 for Poisson and i.i.d. cluster point process frameworks. \label{fig:comp_card}}
	
	\vspace{-10pt}
\end{wrapfigure} 

\begin{figure}[h!]
    \centering
 \includegraphics[width=6.5in,height=6.5in]{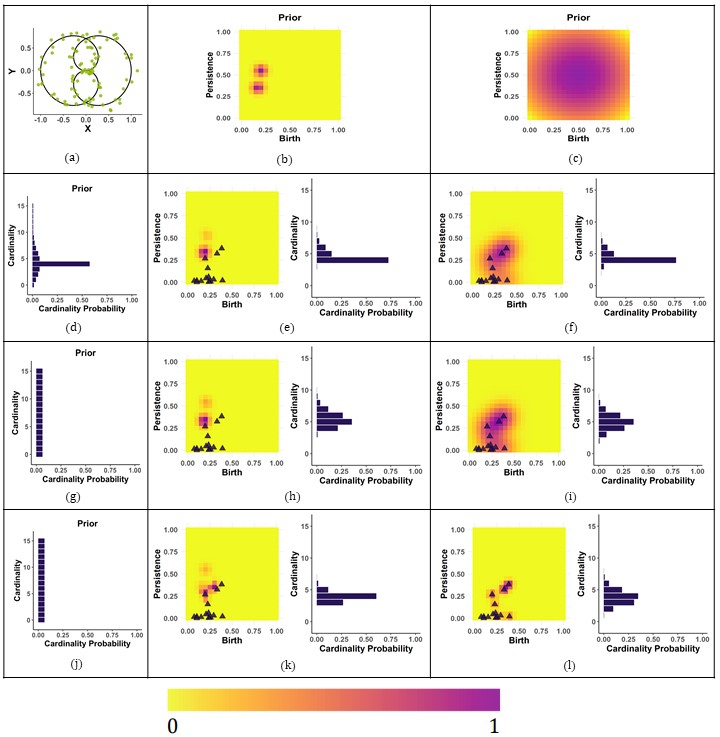}
  \caption{\footnotesize{
  Posterior intensities and cardinalities obtained for Case-2 by using Proposition \ref{prop:post}
  }}

    \label{table:example_post_2}
\end{figure}

\begin{figure}[h!]
    \centering
 \includegraphics[width=6.5in,height=6.5in]{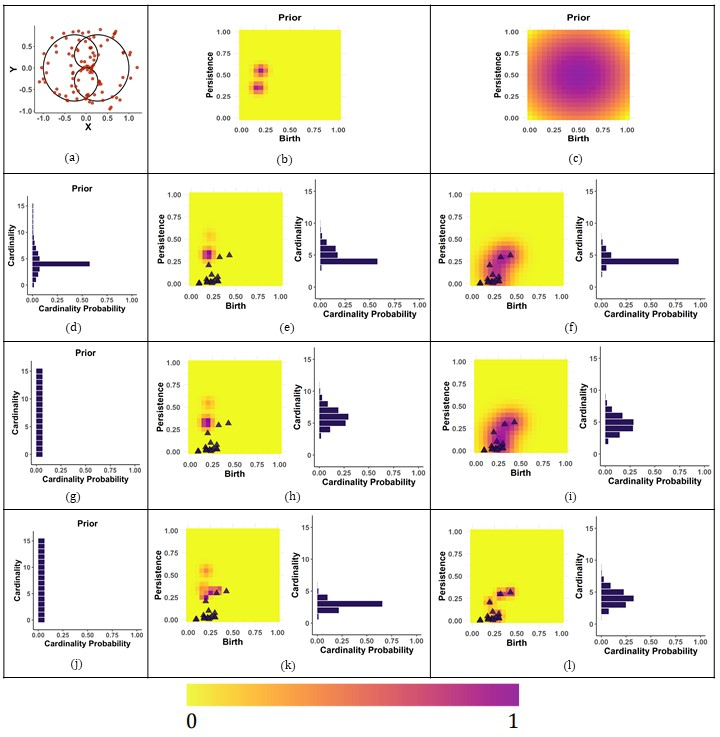}
  \caption{\footnotesize{Posterior intensities and cardinalities obtained for Case-3 by using Proposition \ref{prop:post}
  }}
    \label{table:example_post_3}
\end{figure}

The observed PDs are generated from point clouds sampled uniformly from the polar curve and perturbed by varying levels of Gaussian noise with variances  $0.001I_2$ (Fig. \ref{table:example_post_1} (a)), $0.005I_2$ (Fig. \ref{table:example_post_2} (a)), and $0.01I_2$ (Fig. \ref{table:example_post_3} (a)) which are considered in Case-1, Case-2, and Case-3 respectively. 
Consequently, their PDs exhibit distinctive characteristics such as four prominent features with high persistence and very few spurious features, four prominent features with medium persistence and several spurious features, and three prominent features with medium persistence and many spurious features.

We commence by defining an i.i.d cluster PP with two 
types of prior intensities and cardinalities: (i) informative and (ii) uninformative. 
The prior intensities are
modeled by a Gaussian mixture 
as discussed in 
(M1).
Due to the symmetric nature of the polar curve, in a noiseless scenario, the corresponding PD includes one longer and one shorter persistence point, each with multiplicity $2$. 
Hence we use two  Gaussian components weighted by $2$ for the informative intensity (II) (see Figs. \ref{table:example_post_1}, \ref{table:example_post_2}--\ref{table:example_post_3} (b)). 
To present the intensity maps uniformly throughout this example, we divide the intensities by their corresponding maxima.
This ensures all intensities are on a scale from $0$ to $1$. 
The informative cardinality (IC) is determined by using a discrete distribution with the highest probability at cardinality $4$ (see Figs. \ref{table:example_post_1}, \ref{table:example_post_2}--\ref{table:example_post_3} (d)).  
On the other hand for the uninformative intensity (UI), we use one Gaussian component, and for the uninformative cardinality (UC) we use a discrete uniform distribution (see Figs.
\ref{table:example_post_1}, \ref{table:example_post_2}--\ref{table:example_post_3} (c) and (g) respectively). We present the list of parameters used to define the prior PP in Table \ref{table:prior parameters3}. We examine the cases below.

\noindent \textbf{Case-1:}
 The point cloud considered here is shown in Fig. \ref{table:example_post_1} (a). The 1-dimensional features in the corresponding PD are presented as black triangles overlaid on the posterior intensity plots. 
 We examine the posterior intensity and cardinality for four different combinations of priors - (a) (II, IC), (b) (UI, IC), (c) (II, UC), and (d) (UI, UC).
As the PD consists of a very low number of spurious features, we observe that the posterior computed from any combination of the four predicts the existence and position of all 1-dimensional features accurately (Fig. \ref{table:example_post_1} (e), (f), (h), and (i)).
The uninformative prior cardinality also produces very low variance in the posterior cardinality estimation. 
Since both of the posterior intensity and cardinality estimations are accurate, for the sake of space, we avoid presenting the sensitivity analysis for this case.

Furthermore, for this case we 
 present a comparison between the cardinality statistics  given by using i.i.d. cluster point process characterization of the PD presented herein and a Poisson point process framework presented  in \cite{Maroulas2019a} that estimates the number of homological features by integrating the estimated posterior intensity. As discussed earlier, the  Poisson PP framework approximates the cardinality as a Poisson distribution, and consequently this estimation produces higher variability as the number of points increases. However, the i.i.d. cluster PP characterization leads to accurate estimation of the cardinality with tighter variance (see Fig. \ref{fig:comp_card}).

\vspace{0.1in}
\noindent \textbf{Case-2:} We consider all of the priors as in Case-1. The point cloud used for this case (Fig. \ref{table:example_post_2} (a)) is more perturbed around the polar curve than Case-1 (Gaussian noise with variance $0.005I_2$). 
The associated PD, presented as black triangles overlaid on the posterior intensity plots,  exhibits more spurious features. The parameters used for this case are listed in Table \ref{table:prior parameters2}. 
First, we estimate the posterior intensity and cardinality for all four combinations using the same parameters as in Case-1, and the results are presented in Fig. \ref{table:example_post_2} (e), (f), (h), and (i). 
For the combinations (II, IC)  and (UI, IC) of priors, the posterior intensity and cardinality can accurately estimate the holes with different variance levels. 
However, due to the presence of several spurious features the other two combinations, (II, UC) and (UI, UC), slightly overestimate the cardinality. 
Next, to illustrate the effect of observed data on the posterior, we adjust two parameters, the variance of the likelihood $\sigma_{D_{Y_O}}$ and the decay parameter of the unexpected features, $\mu_{D_{Y_U}}$. 
Recall that the intensity density of the PP $\D_{Y_U}$, consisting of the unexpected features in the observation, is exponential (Eqn. \eqref{eqn:intensity of Dys Gauss}), where $\mu_{D_{Y_U}}$
controls the rate of decay away from the origin. We present the updated   
 posteriors from the two combinations of priors (II, UC)  and (UI, UC). 
By decreasing the variance of the likelihood $\sigma_{D_{Y_O}}$,  the posterior intensities rely more on the observed features in the PD (see Fig. \ref{table:example_post_2} (k) and (l)).
On the other hand, by adjusting the decay parameter, we enable our model to recognize the presence of several spurious features in PD. 
This improves the estimation of posterior cardinality, which is evident in Fig. \ref{table:example_post_2} (k) and (l).

\vspace{0.1in}
\noindent \textbf{Case-3:} In this case we consider the point cloud (Fig. \ref{table:example_post_3} (a)), which is very noisy (Gaussian noise with variance $0.01I_2$). 
Due to the noise level, we encounter only three points with medium prominent persistence, and there are many spurious features.  
All the priors are the same as in Case-1 and Case-2. 
The associated PD is presented as black triangles overlaid on the posterior intensity plots. The parameters used for this case are listed in Table \ref{table:prior parameters2}. 
First, we estimate the posterior intensity and cardinality for all four combinations using the same parameters as in Case-1, and the results are presented in Fig. \ref{table:example_post_3} (e), (f), (h), and (i). 
For the combinations  (II, IC)  and (UI, IC) of priors, the posterior intensity and cardinality can accurately estimate the position and number of 1-dimensional features with different variance levels. 
Due to the presence of several spurious features, the other two combinations (II, UC)  and (UI, UC) overestimate the cardinality distribution. 
Also, in the latter case the posterior intensity estimates the location of the hole with higher variance and is skewed towards the noise features. 
Next, to illustrate the effect of the observed features on the posterior, we adjust two parameters, the variance of the likelihood $\sigma_{D_{Y_O}}$ and the unexpected feature cardinality parameter $\rho_y$  in the posterior estimation for the two combinations (II, UC)  and (UI, UC). 
By decreasing $\sigma_{D_{Y_O}}$ we notice that the posterior intensities rely more on the observed features in the PD (see Fig. \ref{table:example_post_3} (k) and (l)).
On the other hand by increasing  $\rho_y$ the model is able to identify that there are more spurious features in this PD than that of Case-1 and Case-2. 
This improves the estimation of posterior cardinality, which is evident in Fig. \ref{table:example_post_3} (k) and (l).


\section{Classification of Actin Filament Networks} \label{sec:class}
\begin{figure}[h!]
    \centering
       \subfigure[]{\includegraphics[width=1.45in,height=1.45in]{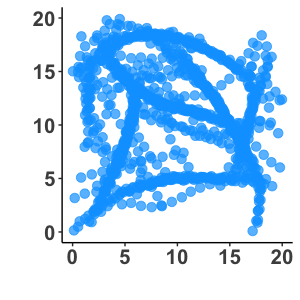}}
    \subfigure[]{\includegraphics[width=1.45in,height=1.45in]{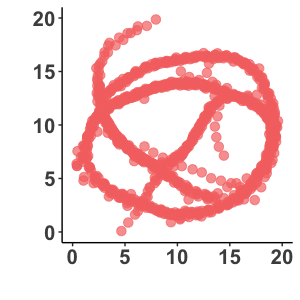}}
     \subfigure[]{\includegraphics[width=1.45in,height=1.45in]{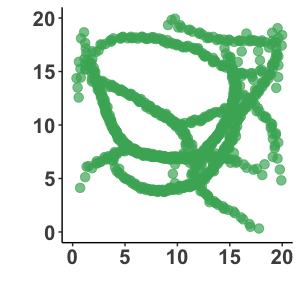}}\\
   \hspace{0.2in} \subfigure[]{\includegraphics[width=1.45in,height=1.45in]{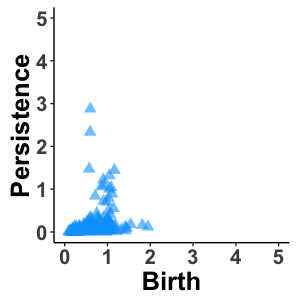}}
    \subfigure[]{\includegraphics[width=1.45in,height=1.45in]{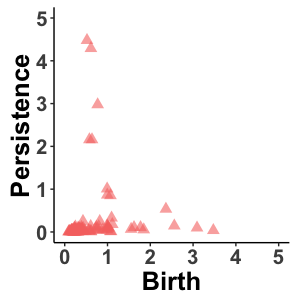}}
     \subfigure[]{\includegraphics[width=1.45in,height=1.45in]{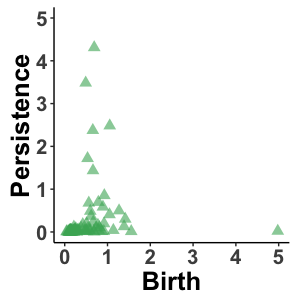}}
    \caption{ (a)--(c) are  examples of  PDs generated from networks in $\mathcal{C}_1, \mathcal{C}_2$, and $\mathcal{C}_3$ respectively. (d)--(f) are their corresponding PDs. } 
    \label{fig:Class_Pds}
\end{figure}

In this section, we classify 150 actin filament networks in plant cells. 
Such filaments are key in the study of intracellular transportation in plant cells, as these bundles and networks make up the actin cytoskeleton, which determines the structure of the cell and enables cellular motion. 
We examine the classification scheme using three classes of filament networks designated by their respective protein binding numbers (see Fig. \ref{fig:Class_Pds} (a)-(c) for examples). 
Higher numbers of cross-linking proteins produce thicker actin cables (\cite{Tang2014}), and in turn, indicate local geometric signatures. 
However, the differences are not always notable due to the presence of noise in the data. 
To bypass this, we explore these networks by means of their respective PDs as they distill salient information about the network patterns with respect to connectedness and empty space (holes), i.e. we can differentiate between filament networks by examining their homological features. 
In particular, we focus on classifying simulated image networks generated at the Abel Research Group with the number of cross-linking proteins $N=825, 1650$, and $3300$, which are denoted as $\mathcal{C}_1, \mathcal{C}_2$, and $\mathcal{C}_3$, respectively. 
The networks consist of the coordinates for the actin filaments and were created using the AFINES stochastic simulation framework introduced in \cite{freedman2018nonequilibrium,freedman2017versatile}, which models the assembly of the actin cytoskeleton. The value of each parameter in the simulation process is chosen to mimic real actin filaments.

From the viewpoint of topology, class $\mathcal{C}_2$ and class $\mathcal{C}_3$ contain more prominent holes than class $\mathcal{C}_1$. Also, their respective PDs have different cardinalities. 
Hence, this topological aspect yields an important contrast between these three classes. 
To capture these differences we employ the following Bayes factor classification approach by relying on the closed form estimation of posterior distributions discussed in Section \ref{sec:gm_post}.  
A PD $D$ that needs to be classified is a sample from an i.i.d. cluster point process $\mathcal{D}$ with  intensity $\lambda_{\mathcal{D}}$ and cardinality $\rho_{\mathcal{D}}$  and its probability density has the form
$ p_{\mathcal{D}}(D)= \rho_{\mathcal{D}}(|D|)\prod_{d \in D}\lambda_{\mathcal{D}}(d).$ For a training set $Q_{Y^k} := D_{Y^k_{1:n}}$ for $k = 1, \cdots, K$ from $K$ classes of random diagrams $\mathcal{D}_{Y^k}$, we obtain the posterior intensities from the Bayesian framework using Proposition \ref{prop:post}.
The posterior probability density of $D$ given the training set $Q_{Y^k}$ is given by
\begin{equation} \label{eqn:poisson_posterior_density}
p_{\mathcal{D}|\mathcal{D}_{Y^k}} (D|Q_{Y^k}) = \rho_{\mathcal{D}|\mathcal{D}_{Y^k}}(|D|)\prod_{d \in D}\lambda_{D|Q_{Y^k}}(d),
\end{equation}
and consequently, the Bayes factor is obtained by the ratio
$BF^{ij}(Q_{Y^i},Q_{Y^j})=\frac{\rho_{D|\mathcal{D}_{Y^i}}(D|Q_{Y^i})}{\rho_{D|\mathcal{D}_{Y^j}}(D|Q_{Y^j})}$ for a class $i,j = 1, \cdots, K$ such that $i \neq j$.
For every pair $(i,j)$, if $BF^{ij}(Q_{Y^i},Q_{Y^j})>c$, we assign one vote to class $Q_{Y^i}$, or otherwise for $BF^{ij}(Q_{Y^i},Q_{Y^j})<c$. 
The final assignment of the class of $D$ is obtained by a majority voting scheme.

\begin{wraptable}{r}{8cm}

\vspace{-0.3in}
	\begin{center}
\footnotesize
		\begin{tabular}  {|l|c|c|c|c|} 
		\hline 	
		\multicolumn{5}{|c|}{\textbf{Parameters for (M1)}}  \\ 
		\hline
		 $\mu_{i}^{\mathcal{D}_X}$ &  $\sigma_{i}^{\mathcal{D}_X}$ & $c_{i}^{\mathcal{D}_X}$& $N_0$ &  $\rho_{\mathcal{D}_{X}}$ \\
		\hline
	$(1,2)$  &  6 &  1 &  25 & $24/25$ \\
		\hline
			\multicolumn{2}{|c|}{\textbf{Parameters for (M2)}} & \multicolumn{3}{|c|}{\textbf{Parameters for (M3)}}  \\ 
		\hline
		 $\sigma^{\mathcal{D}_{Y_O}}$ & $\mu^{\mathcal{D}_{Y_U}}$ &   $M_0$ & $\rho_y$ & $\alpha$  \\
		\hline
  0.01  & 1   &   25 &  $2/25$ & 0.95\\ 
		\hline
	\end{tabular}
	\end{center}
	\vspace{-0.1in}
	\caption{ List of parameters used for the classification.} 	\label{table:class_parameters}
	\vspace{-0.1in}
	\end{wraptable}

PDs with 1-dimensional features (see Fig. \ref{fig:Class_Pds} (d)--(f) for an example of each class)  were created for each actin network through Rips filtration as discussed in Section \ref{subsec:persistence diagram}, which were then used as input for the Bayes factor classification scheme of Eqn. \ref{eqn:poisson_posterior_density}. The number of 1-dimensional features in the dataset is large and the posterior estimation for this dataset is not computationally attainable. To mitigate this issue, we  subsample the dataset to reduce the size of it. Precisely, our subsampled dataset consists of  25 points from each of the PDs obtained from the 150 synthetic filament networks. We  found that taking more than 25 points from each of the PDs did not  improve
the classification, and typically led to a very expensive computational scheme.
The corresponding PDs of these network filaments do not show any discernible pattern, and consequently, we   
adopt a data-driven scheme for classification  using an uninformative flat prior.
Table \ref{table:class_parameters} summarizes the choices of parameters for the model.

\begin{wrapfigure}{r}{0.35\textwidth}
	\vspace{-25pt}
	\centering{
		\includegraphics[width=1.75in,height=1.75in]{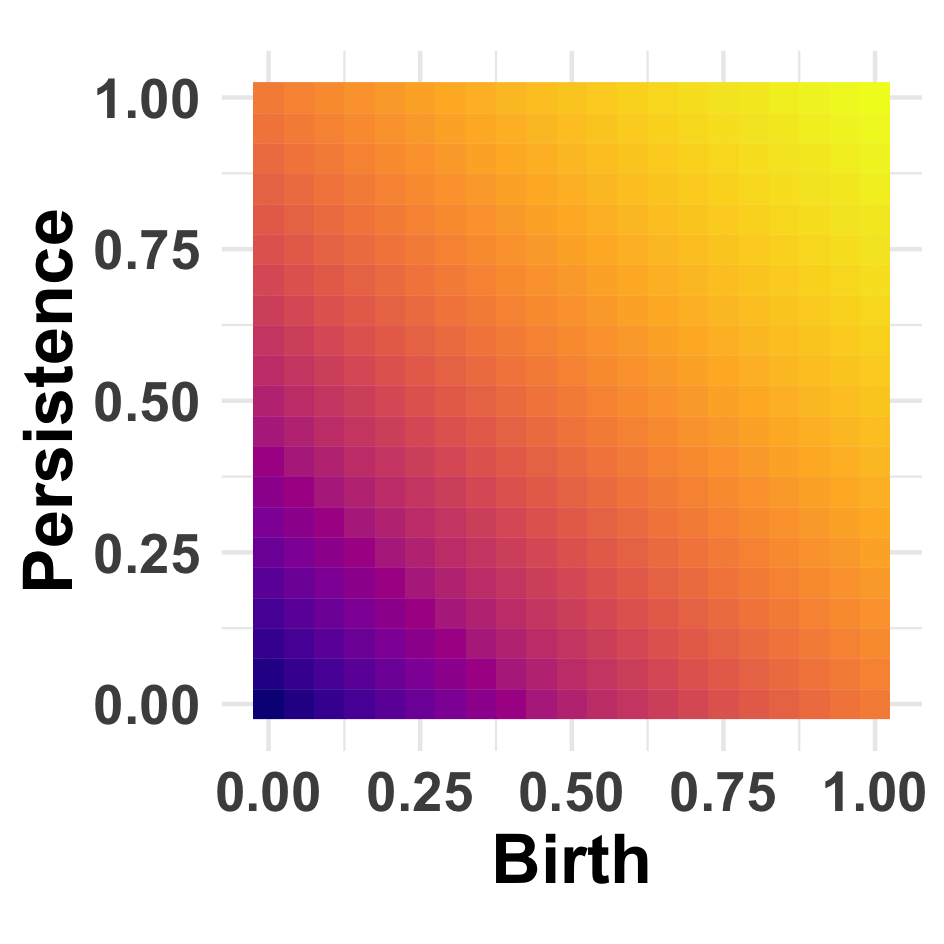}
	}
	\vspace{-10pt}
	\caption{  The intensity density for the unexpected feature PP used in classifying the filament networks. \label{fig:clutter_class}}
	\vspace{-10pt}
\end{wrapfigure} 
One intuitive interpretation of the unexpected features is that they represent the presence of noise in the dataset, consequently they often have very short persistence. 
On the other hand, the dataset of filament networks routinely consists of several incomplete loops (see Fig. \ref{fig:Class_Pds}), which imply that points with late birth and short persistence are expected from the underlying topology. 
Since we use 10-fold cross validation to estimate the model's accuracy, the posterior is calculated using the training set for each fold and each class. 
Then for each instance, we assign the class by using the majority voting scheme.
 We compute the resulting area under the receiver operating characteristic (ROC) curves (AUCs)
and the results are listed in Table \ref{tab:comparison}. The AUC across $10$-folds was 0.925.

{\subsection{Comparison  with Other Methods}
We compared our method with several other machine learning algorithms to benchmark against them. We mainly pursued two avenues - (i) features selected using TDA methodology, and (ii) features selected using non-TDA methodology.  
Two other TDA methods which provide topological summaries and we compare our method with are persistence landscapes (Pls)  \cite{Bubenik2015} and  persistence images (PIs)  \cite{Adams2017}.  
These summaries have been widely implemented as they are amenable to the existing machine learning methodologies. The main theme of these summaries  is the extraction of a pertinent feature vector and implement a classifier trained using machine learning algorithms. Here we input these  topological summaries as features for three different optimized classification algorithms: random forest (RF),  support vector machine (SVM), and  neural network (NN). }

{We considered a vector of 2500 values at which the PLs of order 1, 2, and 3 are evaluated, and found that 
the third order PL to be the most efficient summary for this classification task.
In order to compute the PIs, we discretize the domain space into  a $50 \times 50$ grid with a spread of $0.1$.  The linear ramp function is used to produce weights for computing  PIs. We explore the classification problem using PIs with and without incorporating the linear weights and found that the PIs without any weights provide better accuracy than those with weights. This is justified as the linear ramp function assigns more weights to the higher persistence points leaving the local features to be insignificant. We optimally tune the parameters of SVM using a grid search.  Precisely, the parameter $\gamma$ of the radial basis kernel, that is the inverse of the standard deviation of the kernel,  was optimally selected from a range of $0.1$ to $1$ with a spread of $0.1$. In order to choose the optimal parameters for NN, we performed an extensive grid search for all parameters. However, we found that out of all the parameters, the only two  that can potentially improve the classification accuracy are the number of hidden layers and the maximum number of iterations. The optimal performance was achieved for PLs with 20 layers and maximum iterations of 10  and  for PIs with 3 layers and maximum iterations of 200.  For the RF algorithm we employ 500 trees. }



\begin{figure} [h!]
    \centering
 \subfigure[]{\includegraphics[width=2.5in, height=1.8in]{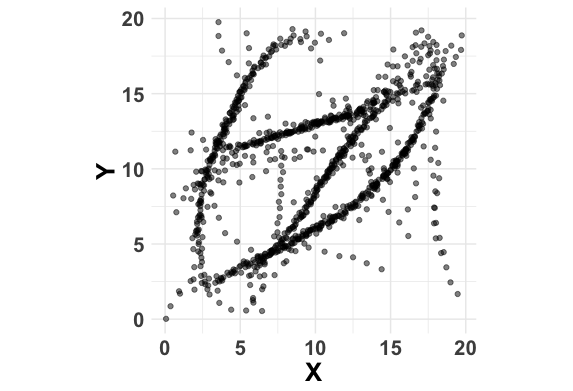}}
  \subfigure[]{\includegraphics[width=2.5in, height=1.8in]{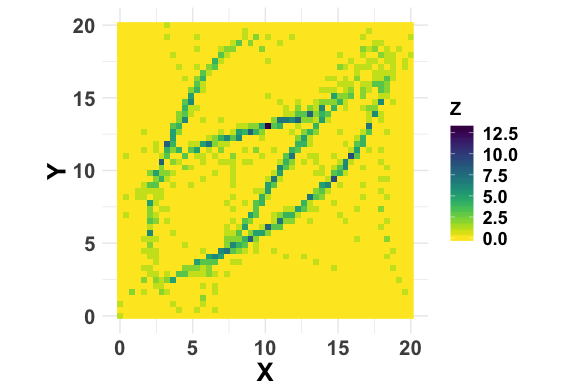}}
  \caption{(a) An example filament network from $\mathcal{C}_1$. (b) The network in (a) converted to a raster image.}
    \label{fig:raster}
\end{figure}

{Additionally, we compare our method with 
machine learning algorithms where the features are selected using a non-TDA method. As the filament networks pose a very definite spatial structure, we found the most useful method to extract the feature is the  Raster images \cite{raster}. 
In particular, the raster image  represents data by using a grid with a value assigned for each pixel. The assigned value can reflect a wide variety of information. In our analysis, we discretize the domain of a filament network into 2500 grid cells identified by 50 rows and 50 columns, and then count the number the points of each grid cell. This approach not only converts each filament network into a raster image which in turn is used as input to  machine learning algorithms but also captures the definite spatial structures such as the presence of empty space and connectedness in a very efficient manner. We present an example in Fig. \ref{fig:raster}. The parameters for the machine learning algorithms are tuned in a similar fashion, i.e., the parameters are optimally tuned using a grid search. The optimal performance for NN was achieved with 5 layers and maximum iterations of 200. The results of this comparison are in Table \ref{tab:comparison}, which showcases that our method outperforms the other methods.}
\begin{table}[h!]
\begin{center}
\begin{tabular}{|c|c|c|c|}
\hline
 \textbf{Method} & \textbf{AUC} & \textbf{Method} & \textbf{AUC}\\
\hline
Bayesian Framework & 0.925 & SVM PL & 0.72  \\
\hline
Random Forest PI & 0.90 & Neural Net PL & 0.79\\ 
\hline
SVM PI & 0.85  & Random Forest Raster & 0.69\\ 
\hline
Neural Net PI & 0.88 & SVM Raster & 0.77 \\
\hline
Random Forest PL &  0.82 & Neural Net Raster & 0.6\\
\hline
\end{tabular}
\end{center}

\caption{{Comparison of methods for filament networks }}\label{tab:comparison}
\end{table}



\section{Conclusion} \label{sec: conclusion}
This paper has proposed a generalized Bayesian framework for PDs 
by modeling them as i.i.d. cluster point processes.
Our framework provides a probabilistic descriptor of the diagrams by simultaneously estimating the cardinality and spatial distributions.
It is noteworthy that our Bayesian model directly employs PDs, which are topological summaries of data, for defining a substitution likelihood rather than using the entire point cloud.  
This deviates from a strict Bayesian model, as we consider the statistics of PDs rather than the underlying datasets used to create them; however, our paradigm incorporates prior knowledge and observed data summaries to create posterior distributions, analogous to the notion of substitution likelihood in \cite{Jeffreys1961}. 
Indeed, the idea of utilizing topological summaries of point clouds in place of the actual point clouds proves to be a powerful tool with applications in wide-ranging fields. 
This process incorporates topological descriptors of point clouds, which simultaneously decipher essential shape peculiarities and avoid unnecessarily complex geometric features.

We derive closed forms of the posterior for realistic implementation, using Gaussian mixtures for the prior intensity and binomials for the prior cardinality.
A detailed example showcases the posterior intensities and cardinalities for various interesting instances created by varying parameters within the model. 
This example exhibits our method's ability to recover the underlying PD.
Thus, the Bayesian inference developed here opens up new avenues for machine learning algorithms and data analysis techniques to be applied \emph{directly} to the space of PDs. 
Indeed, we derive a classification algorithm and successfully apply it to filament network data, while we compare our method with other TDA and machine learning approaches successfully.

\section*{Acknowledgements}
The work has been partially supported by the ARO W911NF-17-1-0313, NSF MCB-1715794 and DMS-1821241, and ARL Co-operative Agreement \# W911NF-19-2-0328. The views and conclusions contained in this document are those of the authors and should not  be  interpreted  as  representing  the  official  policies,  either  expressed  or  implied,of  the  Army  Research  Laboratory  or  the  U.S.  Government.   The  U.S.  Government is authorized to reproduce and distribute reprints for Government purposes not withstanding any copyright notation herein.

\bibliographystyle{apalike}

\bibliography{main}

\newcommand{\noop}[1]{}
\begin{thebibliography}{}

\bibitem[Adams et~al., 2017]{Adams2017}
Adams, H., Emerson, T., Kirby, M., Neville, R., Peterson, C., Shipman, P.,
  Chepushtanova, S., Hanson, E., Motta, F., and Ziegelmeier, L. (2017).
\newblock Persistence images: A stable vector representation of persistent
  homology.
\newblock {\em The Journal of Machine Learning Research}, 18(1):218--252.

\bibitem[Adcock et~al., 2016]{Adcock2016}
Adcock, A., Carlsson, E., and Carlsson, G. (2016).
\newblock The ring of algebraic functions on persistence bar codes.
\newblock {\em Homology, Homotopy and Applications}, 18(1):381--402.

\bibitem[Babichev and Dabaghian, 2017]{Babichev2017}
Babichev, A. and Dabaghian, Y. (2017).
\newblock Persistent memories in transient networks.
\newblock {\em Emergent Complexity from Nonlinearity, in Physics, Engineering
  and the Life Sciences}, 191:179--188.

\bibitem[Bendich et~al., 2016]{Bendich2016}
Bendich, P., Marron, J.~S., Miller, E., Pieloch, A., and Skwerer, S. (2016).
\newblock Persistent homology analysis of brain artery trees.
\newblock {\em The Annals of Applied Statistics}, 10(1):198--218.

\bibitem[Biscio and M{\o{}}ller, 2019]{Biscio2019}
Biscio, C.~A. and M{\o{}}ller, J. (2019).
\newblock The accumulated persistence function, a new useful functional summary
  statistic for topological data analysis, with a view to brain artery trees
  and spatial point process applications.
\newblock {\em Journal of Computational and Graphical Statistics}, pages
  1537--2715.

\bibitem[Bobrowski et~al., 2017]{Bobrowski2017}
Bobrowski, O., Mukherjee, S., and Taylor, J.~E. (2017).
\newblock Topological consistency via kernel estimation.
\newblock {\em Bernoulli}, 23(1):288--328.

\bibitem[Bonis et~al., 2016]{Bonis2016}
Bonis, T., Ovsjanikov, M., Oudot, S., and Chazal, F. (2016).
\newblock Persistence-based pooling for shape pose recognition.
\newblock In {\em Computational Topology in Image Context, ed. A Bac, JL Mari},
  pages 19--29. Springer, New York.

\bibitem[Breuer et~al., 2017]{breuer2017system}
Breuer, D., Nowak, J., Ivakov, A., Somssich, M., Persson, S., and Nikoloski, Z.
  (2017).
\newblock System-wide organization of actin cytoskeleton determines organelle
  transport in hypocotyl plant cells.
\newblock {\em Proceedings of the National Academy of Sciences},
  114(28):E5741--E5749.

\bibitem[Bubenik, 2015]{Bubenik2015}
Bubenik, P. (2015).
\newblock Statistical topological data analysis using persistence landscapes.
\newblock {\em Journal of Machine Learning Research}, 16:77--102.

\bibitem[Bubenik, 2018]{Bubenik2018}
Bubenik, P. (2018).
\newblock The persistence landscape and some of its properties.
\newblock arXiv 1810.04963.

\bibitem[Carlsson, 2009]{Carlsson2009}
Carlsson, G. (2009).
\newblock Topology and data.
\newblock {\em Bulletin of the American Mathematical Society}, 46:255–--308.

\bibitem[Carlsson and de~Silva, 2010]{Carlsson2010}
Carlsson, G. and de~Silva, V. (2010).
\newblock Zigzag persistence.
\newblock {\em Foundations of computational mathematics}, 10(4):367--405.

\bibitem[Carlsson et~al., 2008]{Carlsson2008}
Carlsson, G., Ishkhanov, T., de~Silva, V., and Zomorodian, A. (2008).
\newblock On the local behavior of spaces of natural images.
\newblock {\em International Journal of Computer Vision}, 76(1):1--12.

\bibitem[Carri{\`{e}}re et~al., 2015]{Carriere2015}
Carri{\`{e}}re, M., Oudot, S.~Y., and Ovsjanikov, M. (2015).
\newblock Stable topological signatures for points on {3D} shapes.
\newblock {\em Eurographics}, 34(5):1--12.

\bibitem[Chung et~al., 2015]{Chung2015}
Chung, M.~K., Hanson, J.~L., Ye, J., Davidson, R.~J., and Pollak, S.~D. (2015).
\newblock Persistent homology in sparse regression and its application to brain
  morphometry.
\newblock {\em IEEE Transactions on Medical Imaging}, 34(9):1928--1939.

\bibitem[Ciocanel et~al., 2019]{Ciocanel2019}
Ciocanel, M.-V., Juenemann, R., Dawes, A.~T., and McKinley, S.~A. (2019).
\newblock Topological data analysis approaches to uncovering the timing of ring
  structure onset in filamentous networks.

\bibitem[Daley and Vere-Jones, 1988]{Daley1988}
Daley, D.~J. and Vere-Jones, D. (1988).
\newblock {\em An introduction to the theory of point processes}.
\newblock Springer-Verlag, New York.

\bibitem[Di~Fabio and Ferri, 2015]{Fabio2015}
Di~Fabio, B. and Ferri, M. (2015).
\newblock Comparing persistence diagrams through complex vectors.
\newblock In {\em International Conference on Image Analysis and Processing},
  pages 294--305. Springer.

\bibitem[D{\l{}}otko et~al., 2012]{Dlotko2012}
D{\l{}}otko, P., Ghrist, R., Juda, M., and Mrozek, M. (2012).
\newblock Distributed computation of coverage in sensor networks by homological
  methods.
\newblock {\em Applicable Algebra in Engineering, Communication and Computing},
  23(1--2):29--58.

\bibitem[Edelsbrunner and Harer, 2010]{Edelsbrunner2010}
Edelsbrunner, H. and Harer, J.~L. (2010).
\newblock {\em Computational topology: an introduction.}
\newblock American Mathematical Society, Providence, R.I.

\bibitem[Emmett et~al., 2014]{Emmett2014}
Emmett, K., Rosenbloom, D., Camara, P., and Rabadan, R. (2014).
\newblock Parametric inference using persistence diagrams: a case study in
  population genetics.
\newblock {\em arXiv:1406.4582}.

\bibitem[Fasy et~al., 2014]{Fasy2014}
Fasy, B.~T., Lecci, F., Rinaldo, A., Wasserman, L., Balakrishnan, S., Singh,
  A., et~al. (2014).
\newblock Confidence sets for persistence diagrams.
\newblock {\em The Annals of Statistics}, 42(6):2301--2339.

\bibitem[Freedman et~al., 2017]{freedman2017versatile}
Freedman, S.~L., Banerjee, S., Hocky, G.~M., and Dinner, A.~R. (2017).
\newblock A versatile framework for simulating the dynamic mechanical structure
  of cytoskeletal networks.
\newblock {\em Biophysical journal}, 113(2):448--460.

\bibitem[Freedman et~al., 2018]{freedman2018nonequilibrium}
Freedman, S.~L., Hocky, G.~M., Banerjee, S., and Dinner, A.~R. (2018).
\newblock Nonequilibrium phase diagrams for actomyosin networks.
\newblock {\em Soft matter}, 14(37):7740--7747.

\bibitem[Gameiro et~al., 2015]{Gameiro2015}
Gameiro, M., Hiraoka, Y., Izumi, S., Kramar, M., Mischaikow, K., and Nanda, V.
  (2015).
\newblock A topological measurement of protein compressibility.
\newblock {\em Japan Journal of Industrial and Applied Mathematics},
  32(1):1--17.

\bibitem[Guo et~al., 2018]{Guo2018}
Guo, W., Manohar, K., Brunton, S.~L., and Banerjee, A.~G. (2018).
\newblock Sparse-{TDA}: Sparse realization of topological data analysis for
  multi-way classification.
\newblock {\em IEEE Transactions on Knowledge and Data Engineering}, 30(7):1403
  -- 1408.

\bibitem[Hijmans, 2019]{raster}
Hijmans, R.~J. (2019).
\newblock {\em raster: Geographic Data Analysis and Modeling}.
\newblock R package version 3.0-2.

\bibitem[Humphreys et~al., 2019]{Humphreys2019}
Humphreys, D.~P., McGuirl, M.~R., Miyagi, M., and Blumberg, A.~J. (2019).
\newblock Fast estimation of recombination rates using topological data
  analysis.
\newblock {\em GENETICS}.

\bibitem[Ichinomiya et~al., 2017]{Ichinomiya2017}
Ichinomiya, T., Obayashi, I., and Hiraoka, Y. (2017).
\newblock Persistent homology analysis of craze formation.
\newblock {\em Physical Review E}, 95(1):012504.

\bibitem[Jeffreys, 1961]{Jeffreys1961}
Jeffreys, H. (1961).
\newblock {\em Theory of Probability}.
\newblock Clarendon Press.

\bibitem[Kerber et~al., 2017]{Kerber2017}
Kerber, M., Morozov, D., and Nigmetov, A. (2017).
\newblock Geometry helps to compare persistence diagrams.
\newblock {\em Journal of Experimental Algorithmics (JEA)}, 22:1--4.

\bibitem[Khasawneh and Munch, 2016]{Khasawneh2016}
Khasawneh, F.~A. and Munch, E. (2016).
\newblock Chatter detection in turning using persistent homology.
\newblock {\em Mechanical Systems and Signal Processing}, 70--71:527 -- 541.

\bibitem[Kimura et~al., 2018]{Kimura2018}
Kimura, M., Obayashi, I., Takeichi, Y., Murao, R., and Hiraoka, Y. (2018).
\newblock Non-empirical identification of trigger sites in heterogeneous
  processes using persistent homology.
\newblock {\em Scientific reports}, 8(1):3553.

\bibitem[Kusano et~al., 2016]{Kusano2016}
Kusano, G., Fukumizu, K., and Hiraoka, Y. (2016).
\newblock Persistence weighted {Gaussian} kernel for topological data analysis.
\newblock {\em Proceedings of the 33rd International Conference on Machine
  Learning}, 48:2004--2013.

\bibitem[Lee et~al., 2017]{Lee2017}
Lee, Y., Barthel, S.~D., D{\l}otko, P., Moosavi, S.~M., Hess, K., and Smit, B.
  (2017).
\newblock Quantifying similarity of pore-geometry in nanoporous materials.
\newblock {\em Nature Communications}, 8(1):1--8.

\bibitem[Lum et~al., 2013]{Lum2013}
Lum, P.~Y., Singh, G., Lehman, A., Ishkanov, T., Vejdemo-Johansson, M.,
  Alagappan, M., Carlsson, J., and Carlsson, G. (2013).
\newblock Extracting insights from the shape of complex data using topology.
\newblock {\em Scientific Reports}, 3.

\bibitem[Madison and Nebenf{\"u}hr, 2013]{madison2013understanding}
Madison, S.~L. and Nebenf{\"u}hr, A. (2013).
\newblock Understanding myosin functions in plants: are we there yet?
\newblock {\em Current Opinion in Plant Biology}, 16(6):710--717.

\bibitem[Mahler, 2007]{mahler2007}
Mahler, R. (2007).
\newblock {\em Statistical multisource-multitarget information fusion}.
\newblock Artech House, Boston.

\bibitem[Marchese and Maroulas, 2016]{Marchese2016}
Marchese, A. and Maroulas, V. (2016).
\newblock Topological learning for acoustic signal identification.
\newblock In {\em 2016 19th International Conference on Information Fusion
  (FUSION)}, pages 1377--1381.

\bibitem[Marchese and Maroulas, 2018]{Marchese2018}
Marchese, A. and Maroulas, V. (2018).
\newblock Signal classification with a point process distance on the space of
  persistence diagrams.
\newblock {\em Advances in Data Analysis and Classification}, 12(3):657--682.

\bibitem[Maroulas et~al., 2019]{Maroulas2019}
Maroulas, V., Mike, J.~L., and Oballe, C. (2019).
\newblock Nonparametric estimation of probability density functions of random
  persistence diagrams.
\newblock {\em Journal of Machine Learning Research}, 20(151):1--49.

\bibitem[Maroulas et~al., 2020]{Maroulas2019a}
Maroulas, V., Nasrin, F., and Oballe, C. (2020).
\newblock A {Bayesian} framework for persistent homology.
\newblock {\em SIAM Journal on Mathematics of Data Science}, 2(1):48--74.

\bibitem[Maroulas and Nebenf{\"{u}}hr, 2015]{Marouls2015}
Maroulas, V. and Nebenf{\"{u}}hr, A. (2015).
\newblock Tracking rapid intracellular movements: a {B}ayesian random set
  approach.
\newblock {\em The Annals of Applied Statistics}, 9(2):926--949.

\bibitem[Mike et~al., 2016]{Mike2016}
Mike, J., Sumrall, C.~D., Maroulas, V., and Schwartz, F. (2016).
\newblock Nonlandmark classification in paleobiology: computational geometry as
  a tool for species discrimination.
\newblock {\em Paleobiology}, 42(4):696--706.

\bibitem[Mileyko et~al., 2011]{Mileyko2011}
Mileyko, Y., Mukherjee, S., and Harer, J. (2011).
\newblock Probability measures on the space of persistence diagrams.
\newblock {\em Inverse Problems}, 27(12):124007.

\bibitem[Mlynarczyk and Abel, 2019]{Mlynarczyk2019}
Mlynarczyk, P.~J. and Abel, S.~M. (2019).
\newblock First passage of molecular motors on networks of cytoskeletal
  filaments.
\newblock {\em Phys. Rev. E}, 99:022406.

\bibitem[Moyal, 1962]{Moyal1962}
Moyal, J.~E. (1962).
\newblock The general theory of stochastic population processes.
\newblock {\em Acta Mathematica}, 108(1):1--31.

\bibitem[Nasrin et~al., 2019]{Nasrin2019}
Nasrin, F., Oballe, C., Boothe, D.~L., and Maroulas, V. (2019).
\newblock Bayesian topological learning for brain state classification.
\newblock In {\em Proceedings of 2019 IEEE International Conference on Machine
  Learning and Applications (ICMLA)}.

\bibitem[Nicolau et~al., 2011]{Nicolau2011}
Nicolau, M. M.~P., Levine, A.~J., and Carlsson, G.~E. (2011).
\newblock Topology based data analysis identifies a subgroup of breast cancers
  with a unique mutational profile and excellent survival.
\newblock {\em Proceedings of the National Academy of Sciences},
  108(17):7265--70.

\bibitem[Patrangenaru et~al., 2018]{Patrangenaru2018}
Patrangenaru, V., Bubenik, P., Paige, R.~L., and Osborne, D. (2018).
\newblock Topological data analysis for object data.
\newblock {\em arXiv:1804.10255}.

\bibitem[Perea and Harer, 2015]{Perea2015}
Perea, J.~A. and Harer, J. (2015).
\newblock Sliding windows and persistence: An application of topological
  methods to signal analysis.
\newblock {\em Foundations of Computational Mathematics}, 15(3):799--838.

\bibitem[Pereira and Mello, 2015]{Pereira2015}
Pereira, C. M.~M. and Mello, R.~F. (2015).
\newblock Persistent homology for time series and spatial data clustering.
\newblock {\em Expert Systems with Applications}, 42(15--16):6026--6038.

\bibitem[Porter and Day, 2016]{porter2016filaments}
Porter, K. and Day, B. (2016).
\newblock From filaments to function: the role of the plant actin cytoskeleton
  in pathogen perception, signaling and immunity.
\newblock {\em Journal of integrative plant biology}, 58(4):299--311.

\bibitem[Reininghaus et~al., 2015]{Reininghaus2015}
Reininghaus, J., Huber, S., Bauer, U., and Kwitt, R. (2015).
\newblock A stable multi-scale kernel for topological machine learning.
\newblock In {\em Proceedings of the IEEE conference on computer vision and
  pattern recognition}, pages 4741--4748.

\bibitem[Robinson and Turner, 2017]{Robinson2017}
Robinson, A. and Turner, K. (2017).
\newblock Hypothesis testing for topological data analysis.
\newblock {\em Journal of Applied and Computational Topology}, 1(2):241--261.

\bibitem[Rouse et~al., 2015]{Rouse2015}
Rouse, D., Watkins, A., Porter, D., Harer, J., Bendich, P., Strawn, N., Munch,
  E., DeSena, J., Clarke, J., Gilbert, J., et~al. (2015).
\newblock Feature-aided multiple hypothesis tracking using topological and
  statistical behavior classifiers.
\newblock In {\em Signal processing, sensor/information fusion, and target
  recognition XXIV}, volume 9474, page 94740L. International Society for Optics
  and Photonics.

\bibitem[{Seversky} et~al., 2016]{Seversky2016}
{Seversky}, L.~M., {Davis}, S., and {Berger}, M. (2016).
\newblock On time-series topological data analysis: new data and opportunities.
\newblock In {\em 2016 IEEE Conference on Computer Vision and Pattern
  Recognition Workshops (CVPRW)}, pages 1014--1022.

\bibitem[Sgouralis et~al., 2017]{Sgouralis2017}
Sgouralis, I., Nebenf{\"{u}}hr, A., and Maroulas, V. (2017).
\newblock A {B}ayesian topological framework for the identification and
  reconstruction of subcellular motion.
\newblock {\em SIAM Journal on Imaging Sciences}, 10(2):871--899.

\bibitem[Shimmen and Yokota, 2004]{Shimmen2004}
Shimmen, T. and Yokota, E. (2004).
\newblock Cytoplasmic streaming in plants.
\newblock {\em Curr Opin Cell Biol.}, 16(1):68--72.

\bibitem[Silva and Ghrist, 2006]{Silva2006}
Silva, V.~D. and Ghrist, R. (2006).
\newblock Coordinate-free coverage in sensor networks with controlled
  boundaries via homology.
\newblock {\em Journal of Robotics Research}, 25(12):1205--1222.

\bibitem[Silva and Ghrist, 2007]{Silva2007}
Silva, V.~D. and Ghrist, R. (2007).
\newblock Homological sensor networks.
\newblock {\em Notices of the American mathematical society}, 54(1).

\bibitem[Sizemore et~al., 2018]{Sizemore2018}
Sizemore, A.~E., Phillips-Cremins, J.~E., Ghrist, R., and Bassett, D.~S.
  (2018).
\newblock The importance of the whole: Topological data analysis for the
  network neuroscientist.
\newblock {\em Network Neuroscience}.

\bibitem[Staiger et~al., 2000]{Staiger2000}
Staiger, C., Baluska, F., Volkmann, D., and Barlow, P. (2000).
\newblock {\em Actin: A Dynamic Framework for Multiple Plant Cell Functions}.
\newblock Springer, 1st edition.

\bibitem[Streit, 2013]{Streit2013}
Streit, R. (2013).
\newblock The probability generating functional for finite point processes, and
  its application to the comparison of {PHD} and intensity filters.
\newblock {\em Journal of Advances in Information Fusion}, 8(2):119--132.

\bibitem[Tang et~al., 2014]{Tang2014}
Tang, H., Laporte, D., and Vavylonisa, D. (2014).
\newblock Actin cable distribution and dynamics arising from cross-linking,
  motor pulling, and filament turnover.
\newblock {\em Mol Biol Cell}, 25(19):3006--3016.

\bibitem[Thomas et~al., 2009]{thomas2009actin}
Thomas, C., Tholl, S., Moes, D., Dieterle, M., Papuga, J., Moreau, F., and
  Steinmetz, A. (2009).
\newblock Actin bundling in plants.
\newblock {\em Cell motility and the cytoskeleton}, 66(11):940--957.

\bibitem[Townsend et~al., 2020]{Townsend2020}
Townsend, J., Micucci, C.~P., Hymel, J.~H., Maroulas, V., and Vogiatzis, K.~D.
  (2020).
\newblock Representation of molecular structures with persistent homology for
  machine learning applications in chemistry.
\newblock {\em Nat Commun}, 11:3230.

\bibitem[Turner et~al., 2014]{Turner2014}
Turner, K., Mileyko, Y., Mukherjee, S., and Harer, J. (2014).
\newblock Fr{\'{e}}chet means for distributions of persistence diagrams.
\newblock {\em Discrete and Computational Geometry}, 52(1):44--70.

\bibitem[Venkataraman et~al., 2016]{Venkataraman2016}
Venkataraman, V., Ramamurthy, K.~N., and Turaga, P. (2016).
\newblock Persistent homology of attractors for action recognition.
\newblock In {\em 2016 IEEE International Conference on Image Processing
  (ICIP)}, pages 4150--4154.

\bibitem[Xia et~al., 2014]{Xia2015}
Xia, K., Feng, X., Tong, Y., and Wei, G.~W. (2014).
\newblock Persistent homology for the quantitative prediction of fullerene
  stability.
\newblock {\em Journal of Computational Chemistry}, 36(6):408--422.

\end{thebibliography}

\section{ Appendix}
This section is organized as follows:
\begin{enumerate}
    \item In Subsection \ref{sup_pgfl}, we provide the necessary definitions and  theorems related to the probability generating functional, which will be
heavily used in the proof of Theorem \ref{thm:bayes}.
\item In Subsection \ref{sup_proof}, we provide the proof of our main theorem.
\item In Subsection \ref{sup_prop_proof}, we provide the proof of proposition \ref{prop:post}.
\end{enumerate}

\subsection{Probability Generating Functional \label{sup_pgfl}}

To calculate the posterior distributions for the Bayesian analysis, the probability generating functional (PGFL) is used.
The PGFL is a point process analogue of the probability generating function (PGF) of random variables. 
Intuitively, the point process can be characterized by the functional derivatives of the PGFL (\cite{Moyal1962}).   

\begin{definition}
The elementary symmetric function $e_{K,k}$ is given by
$ e_{K,k}(\nu_1, \cdots, \nu_K) = \\ \sum_{0 \leq i_1 < \cdots < i_k \leq K} \nu_{i_1}\cdots \nu_{i_k}$ 
with $e_{0,k}=1$ by convention. 
	  \end{definition}
	  
\setcounter{equation}{0}
\begin{definition}
 	\label{def:pgfl}
 	Let $\Psi$ be a finite PP on $\X$ and $\mathcal{H}$ be the Banach space of all bounded measurable complex valued functions $\zeta$ on $\X$. 
 	For a symmetric function,  $\zeta(\mathbf{x})=\zeta(x_1)\cdots \zeta(x_n)$ and $\mathbf{x} = (x_1, \cdots, x_n) \in \X$, the PGFL of $\Psi$ is given by
 	\begin{equation}\label{eqn:univariate_pgfl}
 \small	G[\zeta] =  \mathbb{E} \big[\prod_{j=1}^{n} \zeta(x_{j})\big] = \sum_{n=0}^{\infty}\frac{1}{n!}\int_{\mathbb{X}^{n}}\left(\prod_{j=1}^{n} \zeta(x_{j})\right)\mathbb{J}_{n}(dx_{1}\dots dx_{n}).
 	\end{equation}
\end{definition}

\noindent The first expression shows the analogy of the PGFL with the PGF, as it is the expectation of the product $\prod_{j=1}^{n} \zeta(x_{j})$. 
Hence, if $\zeta(x_i) = x$, a constant real  non-negative number for all $x_i$, then $G[\zeta]$ takes the form of a PGF $g_N(x)= \sum_{n=0}^\infty p_N(n) x^n$, where $p_N(n)$ is the probability distribution of a random $N \in \N_0=\{0,1,2,\cdots \}$.
\begin{remark}
\label{remark:pgfl of iid cluster}
 For an i.i.d. cluster process $\Psi$ the PGFL has the  form
(\cite{Daley1988}):
	\begin{equation} \label{eqn:pgfl of iid cluster}
\small	G[\zeta]=g_N\Big(\int_{\mathcal{X}}\zeta(x)f(x)dx\Big),
	\end{equation}
	where $g_N$ is the PGF of the cardinality $N$, $\zeta$ has the same form as in Def. \ref{def:pgfl}, and $f$ is the probability density discussed after Def. \ref{def:iid cluster} in the main paper. 
  \end{remark}

Next, we define the PGFL for bivariate and conditional point processes as they will enable us to formulate necessary measures for the Bayesian framework. 
We consider the bivariate point process $(\Psi, \Psi_M)$ on the product space $(\mathbb{X}\times  \mathbb{M}, \mathcal{X} \times \mathcal{M})$, where $\mathbb{X}$ and $\mathbb{M}$ are Polish spaces, and  $\mathcal{X}$ and $\mathcal{M}$ are their Borel $\sigma$-algebras respectively. 
For a symmetric measurable complex valued function $\eta$ on $\mathcal{M}$, the variate PGFL will be the expectation of the product $\prod_{j=1}^{n} \zeta(x_{j})\prod_{i=1}^{k} \eta(m_{i})$. Consequently, we obtain
\begin{equation}
\label{eqn:bivariate_pgfl}
\small G^{(\Psi,\Psi_M)}[\zeta, \eta] =  \sum_{n \geq 0} \sum_{k \geq 0}\frac{1}{n!}\frac{1}{k!}\int_{\mathbb{X}^{n}}\int_{\mathbb{M}^{k}}\left(\prod_{j=1}^{n} \zeta(x_{j})\prod_{i=1}^{k} \eta(m_{i})\right) \mathbb{J}_{n,k}^{(\Psi,\Psi_M)}(d\mathbf{x},d\mathbf{m}).
\end{equation}

For ease of notation we write $d\mathbf{m} = dm_{1}\dots dm_{k}$ and $d\mathbf{x}=dx_{1}\dots dx_{n}$. 
The marked PP $(\Psi,\Psi_{M})$ as defined in Def. \ref{def:marked_iid cluster_process} in the main paper is a  bivariate PP which is composed of bijections between points of $\mathbb{X}$ and  $\mathbb{M}$. So, the process has a Janossy measure $\mathbb{J}_{n,k}^{\Psi_M|\Psi}$ and according to \cite{Moyal1962} the PGFL of the conditional PP has the following form
\begin{equation}
\label{eqn:conditional_pgfl}
\small G^{(\Psi_M|\Psi)}[\eta|\mathbf{x}] =  \sum_{k \geq 0}\frac{1}{k!}\int_{\mathbb{M}^{k}}\left(\prod_{i=1}^{k} \eta(m_{i}))\right)\mathbb{J}_k^{(\Psi_M |\Psi)}(d\mathbf{m}).
\end{equation}

\noindent The joint Janossy measure of the bivariate point process $(\Psi,\Psi_{M})$ is given by $\mathbb{J}_{n,k}^{(\Psi,\Psi_M)}(d\mathbf{x},d\mathbf{m}) = \mathbb{J}_{k}^{(\Psi_M|\Psi)}(d\mathbf{m})\mathbb{J}_{n}^{\Psi}(d\mathbf{x})$. Hence by substituting this and Eqn. \eqref{eqn:conditional_pgfl} in Eqn. \eqref{eqn:bivariate_pgfl} we obtain
\begin{equation}
\label{eqn:bivariate_pgfl_11}
\small G^{(\Psi,\Psi_M)}[\zeta, \eta] =  \sum_{n \geq 0}\frac{1}{n!}\int_{\mathbb{X}^{n}}\left(\prod_{j=1}^{n} \zeta(x_{j})\right)G^{(\Psi_M|\Psi)}[\eta|\mathbf{x}] \mathbb{J}_n^{\Psi}(d\mathbf{x}).
\end{equation}

\noindent The final set of tools includes the definition and pertinent properties of functional derivatives that allow us to recover the intensity and cardinality of the posterior of the PDs. 
\begin{definition} \label{def:functional_deriv}
	For a PGFL $G$ as in Eq. \eqref{eqn:pgfl of iid cluster},
	the gradient derivative of $G$ in the direction of $\eta$ evaluated at $\zeta$ is given by $\delta G[\zeta;\gamma] = \lim_{\epsilon \rightarrow 0} \frac{G[\zeta + \epsilon\gamma] - G[\zeta]}{\epsilon}$.
	For $\gamma=\delta_x$, the Dirac delta function centered at $x$, the gradient derivative $\delta G[\zeta;x]$ is called the functional derivative in the direction of $x$.
\end{definition}

\begin{remark}
\label{remark:fncl_prod}
	The functional derivative satisfies the product and chain rules (\cite{mahler2007}). For instance, the product rule is given by 
	\begin{equation} \label{product rule for funcional derivative}
\small	\delta G_1.G_2[\zeta;x]=\delta G_1[\zeta;x]G_2[\zeta] +G_1[\zeta]\delta G_2[\zeta;x].
	\end{equation}
	Consequently, for $X=\{x_1, \cdots x_m\}$ and a subset $\widetilde{X}$ of $X$, the general product rule for functional derivatives can be obtained iteratively as
	\begin{equation} \label{general product rule for funcional derivative}
\small	\delta G_1.G_2[\zeta;X]=\sum_{\widetilde{X}}\delta G_1[\zeta;X \setminus\widetilde{X}].\delta G_2[\zeta;\widetilde{X}], 
	\end{equation}
	where $X \setminus\widetilde{X} = \{x \in X \,| \,x \notin \widetilde{X} \}$. Also, for a linear functional  $f[\zeta]=\int_{\mathcal{X}}\zeta(x)f(x)dx$, the functional derivative in the direction of $z$ is given by $\delta f[\zeta;z]=f(z)$.
Using the chain rule, the functional derivative of the PGFL of an i.i.d. cluster PP is given by
	 \begin{equation} \label{functional derivetive of iid cluster pgfl}
\small	 \delta^{(m)} G[\zeta; X]=g_N^{(m)}(f[\zeta])f(x_1)\cdots f(x_m).
	 \end{equation}
	\end{remark}
	
	The following theorem gives the form of the PGFL for a conditional PP and thus for a marked PP. 	The proof can be found in \cite {Streit2013}.
	
	\begin{theorem}[]
		\label{thm:bayes posterior point process}
		Consider the PGFL for a marked process $G^{(X,Y)}[\zeta, \eta]$ in Eqn. \eqref{eqn:bivariate_pgfl}  and a finite PP $Y=\{y_1, \cdots, y_m\} \in \mathcal{M}$. Then the PGFL of the conditional PP $(X|Y)$ is given by
		    \begin{equation}\label{eqn:conditional pgfl}
	\small	    G^{(X|Y)}[\zeta] =  \frac{\delta^{(0,m)} G^{(X,Y)}[\zeta;\varnothing, 0;y_1\cdots y_m]}{\delta^{(0,m)} G^{(X,Y)}[1;\varnothing, 0;y_1\cdots y_m]}, 
		    \end{equation}
		    where $\delta^{(0,m)} G^{(X,Y)}$ represents no  functional derivative of $G$ with respect to the first argument $\zeta$ and the  derivative  with respect to the second argument $\eta$ in m directions $\{y_1, \cdots,y_m\}$.  
	\end{theorem}
	
	\subsection{Proof of Theorem \ref{thm:bayes} \label{sup_proof}}

\begin{proof}
The Theorem states that the PDs $D_{Y_{1:m}} = D_{Y_1}, \cdots, D_{Y_m}$ are independent samples from the PP $\D_Y$ with cardinality $K_1, \cdots, K_m$ respectively. Now for independent and identical copies $\D_X^i$ of the i.i.d. cluster point process $\D_X$, we have intensity $\lambda_{\D_X} = \frac{1}{m}\sum_{i=1}^{m}\lambda_{\D_{X^i}}$ and cardinality $\rho_{\D_X} = \frac{1}{m}\sum_{i=1}^{m}\rho_{\D_{X^i}}$. Hence without loss of generality, 
\begin{equation}\label{eqn:average_intensity}
  \lambda_{{\D_X}|{D_{Y^{1:m}}}} = \frac{1}{m}\sum_{i=1}^{m}\lambda_{\mathcal{D}_{X^i}|D_{Y^i}} \,\,\,\,  \text{and} \,\,\,\, \rho_{{\D_X}|{D_{Y^{1:m}}}} = \frac{1}{m}\sum_{i=1}^{m}\rho_{\mathcal{D}_{X^i}|D_{Y^i}}.
 \end{equation}
So it is sufficient to compute $\lambda_{\mathcal{D}_{X^i}|D_{Y^i}}$ and $\rho_{\mathcal{D}_{X^i}|D_{Y^i}}$ for fixed $i$. From Eqn. \eqref{eqn:bivariate_pgfl} we have,
\begin{align}
 G^{(\D_{X^i},D_{Y_i})} &= \sum_{K_i,n \geq 0}\frac{1}{K_i!n!}\int_{\mathbb{X}^{n}}\int_{\mathbb{M}^{K_i}}\left(\prod_{l=1}^{K_i} \eta(y_{l})\right)\left(\prod_{j=1}^{n} \zeta(x_{j})\right)\mathbb{J}_{K_i}^{D_{Y_{1:m}}|\D_X}(d\mathbf{y}) \mathbb{J}_n^{\D_X}(d\mathbf{x}) \nonumber\\
 &=  \sum_{n \geq 0}\frac{1}{n!}\int_{\mathbb{X}^{n}}\left(\prod_{j=1}^{n} \zeta(x_{j})\right) \widetilde{G}[\eta|\D_X] \mathbb{J}_n^{\D_X}(d\mathbf{x}).\label{eqn:bivariate_pgfl_1}
 \end{align}
 The second expression is achieved by using the PGFL with respect to $\mathbb{J}_{K_i}^{D_{Y_i}|\D_X}$ obtained by Eqn. \eqref{eqn:conditional_pgfl}. Now, to understand the conditional PP $D_{Y_i}|\D_X$ we need to consider the augmented space  $\W' = \W \cup \{\Delta\}$ where $\Delta$ is a dummy set that will be used for labeling points in $\mathcal{D}_{Y_U}$ (\cite{Maroulas2019a}). Therefore the random set 
$\mathcal{H} = \{(x,y) \in (\mathcal{D}_{X_O},\mathcal{D}_{Y_O})\}\cup\{(\Delta,y)\,\,|\, y \in \mathcal{D}_{Y_U}\}$
 is a marked i.i.d. cluster PP on $\W'\times \W$. The independence condition in Def. \ref{def:marked_iid cluster_process} for marks in $\W$ thus leads to $\widetilde{G}[\eta|\D_X] = \widetilde{G}[\eta|x_1]\cdots \widetilde{G}[\eta|x_n] \widetilde{G}[\eta|\Delta].$ As $\mathcal{D}_{Y_U}$ is an i.i.d. cluster PP and has no association with $\mathcal{D}_{X}$, from Eqn. \eqref{eqn:pgfl of iid cluster}  we get $\widetilde{G}[\eta|\Delta] = S(f_{\mathcal{D}_{Y_U}}[\eta]) $, where $S$ is the PGF of the cardinality distribution of $\mathcal{D}_{Y_U}$. To be consistent with the probability $\alpha(x)$ defined earlier, our Bayesian model deals with two scenarios: either a feature $x$ will not appear in $D_{Y_i}$ with probability $(1-\alpha(x))$ or each $D_{Y_i}$  contains draws from $\ell(y|x)$ associated to a single sample $x$ of $\mathcal{D}_X$ with probability $\alpha(x)$.  Also, by using the fact that Janossy densities $j_1(x)=\ell(x|y_{i})$ and $j_n = 0$ for $n \neq 1$ and using the linearity of the integral, we get $\widetilde{G}[\eta|x_j] = 1-\alpha(x_j)+\alpha(x_j)\int_{\mathbb{M}} \eta(y) \ell (y|x_j)dy$.  Hence Eqn. \eqref{eqn:bivariate_pgfl_1} leads to
 \begin{equation} \label{eqn:bivariate_pgfl_2}
     G[\eta,\zeta] = S(f_{\mathcal{D}_{Y_U}}[\eta])L(\lambda_{\mathcal{D}_{X}}[\zeta(1-\alpha+\alpha\ell_g)]).
 \end{equation}
Here, we denote $\ell_g(x_j)= \int_{\mathcal{M}} \eta(y)\ell(y|x_j)dy$ and $\lambda_{\mathcal{D}_{X}}[\zeta]=\int_{\X}\zeta(x)\lambda_{\mathcal{D}_{X}}(x)dx$ for simplicity of notation. Also, $L$ is the PGF associated to the PGFL $G[\zeta(1-\alpha+\alpha\ell_g)]$. 
Notice that we have the PGFL $G$ as a product of two PGFs. 
This format helps us to find the functional derivatives in an efficient way so that we obtain the PGFL of the conditional PP $\D_X|D_{Y_i}$ as in Eqn. \eqref{eqn:conditional pgfl}. Hence, by using linearity of integral and chain rule of functional derivatives (Eqn. \eqref{general product rule for funcional derivative}), we obtain
\begin{equation}\label{eqn: conditional pfgl_1}
  G^{(\D_X|D_{Y_i})}[\zeta]=\frac{ \sum_{k=0}^{K_i} S^{(K_i-k)}(0). L^{(k)}(\lambda_{\mathcal{D}_{X}}[\zeta(1-\alpha)]).
  	 e_{K_i,k}\big(\frac{\lambda_{\mathcal{D}_X}[ \zeta\alpha \ell (y_1|x)]}{\lambda_{\mathcal{D}_{Y_U}}(y_1)} \cdots \frac{\lambda_{\mathcal{D}_X}[ \zeta\alpha \ell (y_{K_i}|x)]}{\lambda_{\mathcal{D}_{Y_U}}(y_{K_i})}\big)}{ \sum_{k=0}^{K_i} S^{(K_i-k)}(0). L^{(k)}(\lambda_{\mathcal{D}_{X}}[1-\alpha]).e_{K_i,k}\big(\frac{\lambda_{\mathcal{D}_X}[ \alpha \ell (y_1|x)]}{\lambda_{\mathcal{D}_{Y_U}}(y_1)} \cdots \frac{\lambda_{\mathcal{D}_X}[ \alpha \ell (y_{K_i}|x)]}{\lambda_{\mathcal{D}_{Y_U}}(y_{K_i})}\big)}.
 \end{equation}
 For simplifying notation we denote by $e_{K_i,k}(D_{Y_i}) = e_{K_i,k}\big(\frac{\lambda_{\mathcal{D}_X}[ \alpha \ell (y_1|x)]}{\lambda_{\mathcal{D}_{Y_U}}(y_1)} \cdots \frac{\lambda_{\mathcal{D}_X}[ \alpha \ell (y_{K_i}|x)]}{\lambda_{\mathcal{D}_{Y_U}}(y_{K_i})}\big)$ the elementary symmetric function. If $\zeta \equiv z$ is a constant function, then we obtain the PGF of the posterior cardinality distribution as 
 \begin{equation} \label{eqn: conditional pgf_1}
 G^{(\D_X|D_{Y_{i}})}(z)=\frac{\sum_{k=0}^{K_i} z^k S^{(K_i-k)}(0). L^{(k)}(z \lambda_{\mathcal{D}_{X}}[1-\alpha]).
 	e_{K_i,k}(D_{Y_{i}})} { \sum_{k=0}^{K_i} S^{(K_i-k)}(0). L^{(k)}(\lambda_{\mathcal{D}_{X}}[1-\alpha]).e_{K_i,k}(D_{Y_{i}})}.
 \end{equation}

 We derive the cardinality expression first by utilizing the well-known property of the PGF that the probability distribution can be recovered by means of derivatives and by applying the product rule in \eqref{general product rule for funcional derivative} for acquiring the n-th derivative as
 \begin{equation} \label{eqn:post_card}
 	\rho_{\D_{X^i}|D_{Y^i}}(n)=\frac{\sum_{k=0}^{K_i}  S^{(K_i-k)}(0). \frac{1}{(n-k)!}L^{(k)(n-k)}(0).(\lambda_{\mathcal{D}_{X}}[1-\alpha])^{n-k}
 	e_{K_i,k}(D_{Y_{i}})}{ \sum_{k=0}^{K_i} S^{(K_i-k)}(0). L^{(k)}(\lambda_{\mathcal{D}_{X}}[1-\alpha]).	e_{K_i,k}(D_{Y_{i}})}.
 	\end{equation}
 As $S$ and $L$ are the PGFs of the number of points in $\D_{Y_{U}}$ and $\D_X$ respectively, by utilizing well-known properties of the PGF we can write
 \begin{equation}\label{pgf properties}
 S^{(i)}(0)=i!\rho_{\D_{Y_{U}}}(i)\,\,\,\, \text{and} \,\,\,\,L^{(i)}(x)=\sum_{k=i}^{\infty}\,P_i^{k}\,\rho_{\D_{X}}(k).x^{k-i}, 
 \end{equation}  
 where $P$ is the permutation coefficient. Elementary computation thus leads Eqn. \eqref{eqn:post_card} to the desired form of posterior cardinality as in Eqn. \eqref{post cardinality}.
 
 As is proved in \cite{Moyal1962}, the intensity density $\lambda$ of a PP can be obtained by differentiating the corresponding probability generating functional $G$, i.e., $\lambda(x) = \delta G[1;x]$, where $\delta G[1;x]$ is the functional derivative in the direction of $x$ (see Def. \ref{def:functional_deriv}). Generally speaking, one obtains the intensity for a general PP through $\lambda(x) = \lim_{h \rightarrow 1} \delta G[h;x]$, but the preceding identity suffices for our purposes since we only consider PPs for which Eqn. \eqref{eqn: conditional pfgl_1} is defined for all bounded $h$. Hence, we find the required derivative of Eqn. \eqref{eqn: conditional pfgl_1} as  
 
 \vspace{-0.2in}
 \begin{equation}
 \delta G^{(\D_X|D_{Y_{1:m}})}[1;x] =\sum_{i=1}^m\Bigg[\frac{ \sum_{k=0}^{K_i} S^{(K_i-k)}(0). L^{(k+1)}(\lambda_{\mathcal{D}_{X}}[1-\alpha]).
 	E_{K_i,k}(D_{Y_{i}})}{ \sum_{k=0}^K S^{(K_i-k)}(0). L^{(k)}(\lambda_{\mathcal{D}_{X}}[1-\alpha]).E_{K_i,k}(D_{Y_{i}})}(1-\alpha(x))\lambda_{\mathcal{D}_X}(x)  \nonumber
 \end{equation}
 \begin{equation}\nonumber
  +\!\!\!\sum_{y \in D_{Y_{i}}}\!\!\!\! \frac{\alpha(x)\ell(y|x)\lambda_{\mathcal{D}_X}(x)}{\lambda_{\D_{Y_U}}(y)} \,\frac{ \sum_{k=0}^{K_i-1} S^{(K_i-k-1)}(0). L^{(k+1)}(\lambda_{\mathcal{D}_{X}}[1-\alpha]).
 	E_{K_i-1,k}\big(D_{Y_{i}}\big)}{ \sum_{k=1}^{K_i} S^{(K_i-k)}(0). L^{(k)}(\lambda_{\mathcal{D}_{X}}[1-\alpha]).E_{K_i,k}(D_{Y_{i}})	}\Bigg] 
 \end{equation}
 Similarly, 
 using Eqn. \eqref{pgf properties} gives the format of the posterior intensity $\lambda_{\D_X|D_{Y_{1:m}}}$ and this completes the proof.
\end{proof}

\subsection{Proof of Proposition \ref{prop:post} \label{sup_prop_proof}}
\begin{lemma} \label{lemma:prop of gaussian dist}
Let $\mathbf{H}, \mathbf{R}, \mathbf{P}$ be $p \times p$ matrices, $\mathbf{m}$ and $\mathbf{d}$ be $p \times 1$ vectors, and  $\mathbf{R}$ and $\mathbf{P}$ be positive definite. Then $\int\, \mathcal{N}(\mathbf{y};\mathbf{H} \mathbf{x}+\mathbf{d},\mathbf{R})\,\mathcal{N}(\mathbf{x};\mathbf{m},\mathbf{P}) dx=\mathcal{N}(\mathbf{y};\mathbf{H} \mathbf{m}+\mathbf{d},\mathbf{R}+\mathbf{H} \mathbf{P} \mathbf{H}^T)$.
 \end{lemma}

 \begin{lemma} \label{lemma:prop of gaussian dist_1}
Let $\mathbf{H}, \mathbf{R}, \mathbf{P}$ be $p \times p$ matrices, $\mathbf{m}$ be a $p \times 1$ vector, and suppose that $\mathbf{R}$ and $\mathbf{P}$ are positive definite. Then $\mathcal{N}(\mathbf{y};\mathbf{H} \mathbf{x},\mathbf{R})\,\mathcal{N}(\mathbf{x};\mathbf{m},\mathbf{P})=q(\mathbf{y})\,\mathcal{N}(\mathbf{x};\hat{\mathbf{m}},\hat{\mathbf{P}})$, where $q(\mathbf{y})=\mathcal{N}(\mathbf{y};\mathbf{H}\mathbf{m},\mathbf{R}+\mathbf{H} \mathbf{P} \mathbf{H}^T), \,\, \hat{\mathbf{m}}=\mathbf{m}+\mathbf{K}(\mathbf{y}-\mathbf{H}\mathbf{m}), \,\, \hat{\mathbf{P}}=(\mathbf{I} -\mathbf{K} \mathbf{H}) \mathbf{P}$ and $\mathbf{K}=\mathbf{P}\mathbf{H}^T(\mathbf{H}\mathbf{P}\mathbf{H}^T+\mathbf{R})^{-1}$.
 \end{lemma}

\begin{proof}[\textbf{Proof of Proposition \ref{prop:post}}]
The proposition is established by substituting Eqn. \eqref{eqn:intensity and cardinality of Dx} --\eqref{eqn:cardinality of Dys} in Eqn. \eqref{post intensity_1} and \eqref{post cardinality}. This produces an integral involving the product of two Gaussians in the arguments of the elementary symmetric function  $\frac{\lambda_{\mathcal{D}_X}[ \alpha \ell (y|x)]}{ \lambda_{\mathcal{D}_{Y_U}}(y)}$, and we derive this by using Lemma \ref{lemma:prop of gaussian dist}. In particular, note that if  $\mathbf{H} =\mathbf{I}, \mathbf{R} = \sigma^{\mathcal{D}_{Y_O}}\mathbf{I}, \mathbf{m} = \mu_{l}^{\mathcal{D}_{X}},$ and $\mathbf{P} = \sigma_{l}^{\mathcal{D}_{X}}\mathbf{I}$, we write
 \begin{equation} \nonumber
      \alpha \int_{\W} \lambda_{\mathcal{D}_X} (x) \ell (y_i|x) dx = \alpha \sum_{l=1}^N c_l^{\mathcal{D}_X} \mathcal{N} (y;\mu_l^{\mathcal{D}_X}, (\sigma^{\mathcal{D}_{Y_O}}+\sigma_l^{\mathcal{D}_{X}}))= \alpha \langle c^{\mathcal{D}_X}, q (y_i) \rangle .
 \end{equation}
 
The only other portion of the formula that is not immediate is the term $\ell(y|x) \lambda_{\mathcal{D}_{X}}$ in Eqn. \eqref{post intensity_1}, as it is a product of two pertinent Gaussians.
Using Lemma \ref{lemma:prop of gaussian dist_1} with $\mathbf{H} =\mathbf{I}, \mathbf{R} = \sigma^{\mathcal{D}_{Y_O}}\mathbf{I}, \mathbf{m} = \mu_{l}^{\mathcal{D}_{X}},$ and $\mathbf{P} = \sigma_{l}^{\mathcal{D}_{X}}\mathbf{I}$, we have that $\ell(y|x) \lambda_{\mathcal{D}_{X}} = \sum_{l=1}^N c_{l}^{\mathcal{D}_{X}} q_l(y) \mathcal{N}^{*}(x; \mu_{l}^{x|y},\sigma_{l}^{x|y}\mathbf{I})$, with $\mu_{l}^{x|y} = \frac{\sigma_{l}^{\mathcal{D}_X} y+\sigma^{\mathcal{D}_{Y_O}}\mu_{l}^{\mathcal{D}_X}}{\sigma_{l}^{\mathcal{D}_X}+\sigma^{\mathcal{D}_{Y_O}}}; \,\,\,\,\, \text{and}\,\,\,\,	\sigma_{l}^{x|y}= \frac{\sigma^{\mathcal{D}_{Y_O}}\,\sigma_{l}^{\mathcal{D}_X}}{\sigma_{l}^{\mathcal{D}_X}+\sigma^{\mathcal{D}_{Y_O}}}$ as required for $C_l^{x|y}$. 
\end{proof}
\end{document}